%% file: arxiv.tex
\pdfoutput=1
\documentclass[twoside,10.5pt]{article}
\usepackage{graphicx,amsmath,amsfonts,amssymb,color}
\usepackage{multirow}
\usepackage{geometry}
\usepackage{algorithm}
\usepackage{algorithmicx} 
\usepackage{algpseudocode}
\usepackage{caption}
\usepackage{subcaption}
\usepackage[amsmath,amsthm,thmmarks]{ntheorem}
\usepackage{mathtools}
\usepackage{enumitem}
\usepackage{listings}
\usepackage{placeins}
\usepackage{framed}
\usepackage[utf8]{inputenc}
\usepackage[english]{babel}
\usepackage{csquotes}
\usepackage[backend=bibtex,style=numeric,citestyle=numeric-comp,hyperref=true,backref=true]{biblatex}
\input{./tex/macros.tex}

\title{\LARGE Quantifying the probable approximation error of probabilistic inference programs}

\author{
       \large  Marco F. Cusumano-Towner\\
       \footnotesize Computer Science \& Artificial Intelligence Laboratory\\
       \small   Massachusetts Institute of Technology\\
       \small \texttt{marcoct@mit.edu} \\
       \and
       \large  Vikash K. Mansinghka \\
       \footnotesize Department of Brain \& Cognitive Sciences\\
       \small  Massachusetts Institute of Technology\\
       \small  \texttt{vkm@mit.edu} \\
       }

\begin{document}
\date{}
\maketitle

\begin{abstract}
\input{./tex/abstract.tex}

\end{abstract}

\input{./tex/introduction.tex}

\FloatBarrier
\input{./tex/full_page_figure.tex}
\FloatBarrier

\input{./tex/estimating.tex}
\input{./tex/analyzing.tex}
\input{./tex/applications.tex}
\input{./tex/discussion.tex}

\section*{Acknowledgments}
The authors would like to thank Ulrich Schaechtle and Anthony Lu for testing
the technique, and David Wingate, Alexey Radul, Feras Saad, and Taylor Campbell
for helpful feedback and discussions.
This research was supported by DARPA (PPAML program, contract number FA8750-14-2-0004), IARPA (under research contract 2015-15061000003), the Office of Naval Research (under research contract N000141310333), the Army Research Office (under agreement number W911NF-13-1-0212), and gifts from Analog Devices and Google.
\printbibliography

\newpage
\appendix
\input{./tex/basic_notation.tex}
\input{./tex/deriving.tex}

\input{./tex/proofs.tex}

\input{./tex/metainference_analysis.tex}
\input{./tex/application_derivations.tex}

\end{document}

%% file: tex/macros.tex
\newcommand{\E}{\mathrm{E}}
\newcommand{\Var}{\mathrm{Var}}
\newcommand{\KL}{\mathrm{D}_{\mathrm{KL}}}
\newcommand{\SBJ}{\mathrm{D}_{\mathrm{SBJ}}}
\newcommand{\QIS}{\hat{q}_{\mathrm{IS}}}
\newcommand{\QHM}{\hat{q}_{\mathrm{HM}}}
\newcommand*\Let[2]{\State #1 $\gets$ #2}
\newcommand*\Sample[2]{\State #1 $\sim$ #2}

\makeatletter
\newcommand{\subalign}[1]{%
  \vcenter{%
    \Let@ \restore@math@cr \default@tag
    \baselineskip\fontdimen10 \scriptfont\tw@
    \advance\baselineskip\fontdimen12 \scriptfont\tw@
    \lineskip\thr@@\fontdimen8 \scriptfont\thr@@
    \lineskiplimit\lineskip
    \ialign{\hfil$\m@th\scriptstyle##$&$\m@th\scriptstyle{}##$\crcr
      #1\crcr
    }%
  }
}
\makeatother

\newtheorem{frm-def}{Definition}
\newtheorem{frm-prop}{Proposition}
\newtheorem{frm-lemma}{Lemma}
\newtheorem{frm-def-sup}{Definition}
\newtheorem{frm-prop-sup}{Proposition}
\newtheorem{frm-lemma-sup}{Lemma}

%% file: tex/abstract.tex
This paper introduces a new technique for quantifying the approximation error
of a broad class of probabilistic inference programs, including ones based on
both variational and Monte Carlo approaches. The key idea is to derive a
subjective bound on the symmetrized KL divergence between the distribution
achieved by an approximate inference program and its true target distribution.
The bound’s validity (and subjectivity) rests on the accuracy of two auxiliary
probabilistic programs: (i) a ``reference'' inference program that defines a gold
standard of accuracy and (ii) a ``meta-inference'' program that answers the
question ``what internal random choices did the original approximate inference program
probably make given that it produced a particular result?'' The paper includes
empirical results on inference problems drawn from linear regression, Dirichlet
process mixture modeling, HMMs, and Bayesian networks. The experiments show
that the technique is robust to the quality of the reference inference program
and that it can detect implementation bugs that are not apparent from
predictive performance.

%% file: tex/introduction.tex
\section{Introduction}
A key challenge for practitioners of probabilistic modeling is the
approximation error introduced by variational and Monte Carlo inference
techniques. The Kullback-Leibler (KL) divergence \cite{cover2012elements} between the result of
approximate inference — i.e. the variational approximation, or the distribution
induced by one run of the sampler — and the true target distribution is
typically unknown. Predictive performance on a held-out test set is sometimes
used as a proxy, but this need not track posterior convergence.

This paper introduces a new technique for quantifying the approximation error
of a broad class of probabilistic inference programs, including 
variational and Monte Carlo approaches. The key idea is to derive a
``subjective'' bound on the symmetrized KL divergence between the distribution
achieved by an approximate inference program and its true target distribution.
The bound’s validity (and subjectivity) rests on beliefs about the accuracy of
auxiliary probabilistic program(s). The first is a ``reference'' inference
program that defines a gold standard of accuracy but that might be difficult to
compute.  When the original approximate inference program has a tractable
output probability density, this is sufficient. If the output density of the
approximate inference program is not available, then the technique also depends
on the accuracy of a ``meta-inference'' program that answers the question
``what internal random choices did the approximate inference program of
interest probably make, assuming that it produced a particular result that was
actually produced by the reference?'' In Section~\ref{sec:related} we relate
this technique to some recent work.

The technique is implemented as a probabilistic meta-program for the Venture
probabilistic programming platform \cite{Mansinghka2014}, written in the
VentureScript language. The paper includes empirical results on inference
problems drawn from linear regression, Dirichlet process mixture modeling,
HMMs, and Bayesian networks. The experiments show that the technique is robust
to the quality of the reference inference program and that it can detect
implementation bugs that are not apparent from predictive performance.

%% file: tex/full_page_figure.tex
\renewcommand{\ttdefault}{pcr}
\lstset{ %
  backgroundcolor=\color{white},   
  basicstyle=\ttfamily\scriptsize,        
  literate={~} {$\sim$}{1}
}
\begin{figure}[t]
\centering
\begin{subfigure}[b]{1.0\textwidth}
    \centering
    \begin{equation*}
    \resizebox{1.0\textwidth}{!}{
        \setlength{\fboxsep}{1.5\fboxsep}\boxed{
        \SBJ(q(z;x^*)||p(z|x^*)) \ge \KL(q(z;x^*)||p(z|x^*)) + \KL(p(z|x^*)||q(z;x^*)) \;\; \mbox{for} \;\; \KL(r(z;x^*)||p(z|x^*)) = 0}}
    \end{equation*}
\vspace{-5mm}
    \caption{A relationship between subjective divergence $\SBJ$ and symmetrized KL divergence}
\end{subfigure}
    \begin{subfigure}[b]{0.5\textwidth}
        \centering
        \includegraphics[width=1.0\textwidth]{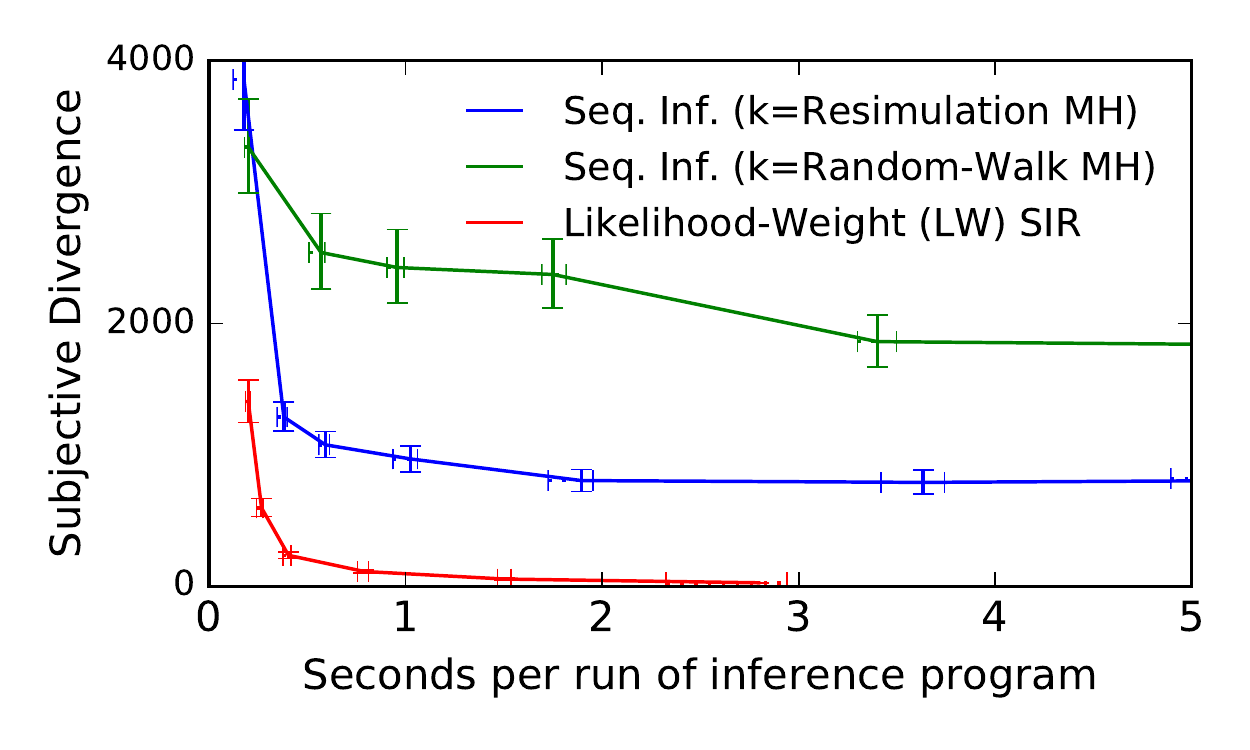}
\vspace{-7mm}
        \caption{}
        \label{fig:linreg_sampling}
    \end{subfigure}\hfill
    \begin{subfigure}[b]{0.5\textwidth}
        \centering
        \includegraphics[width=1.0\textwidth]{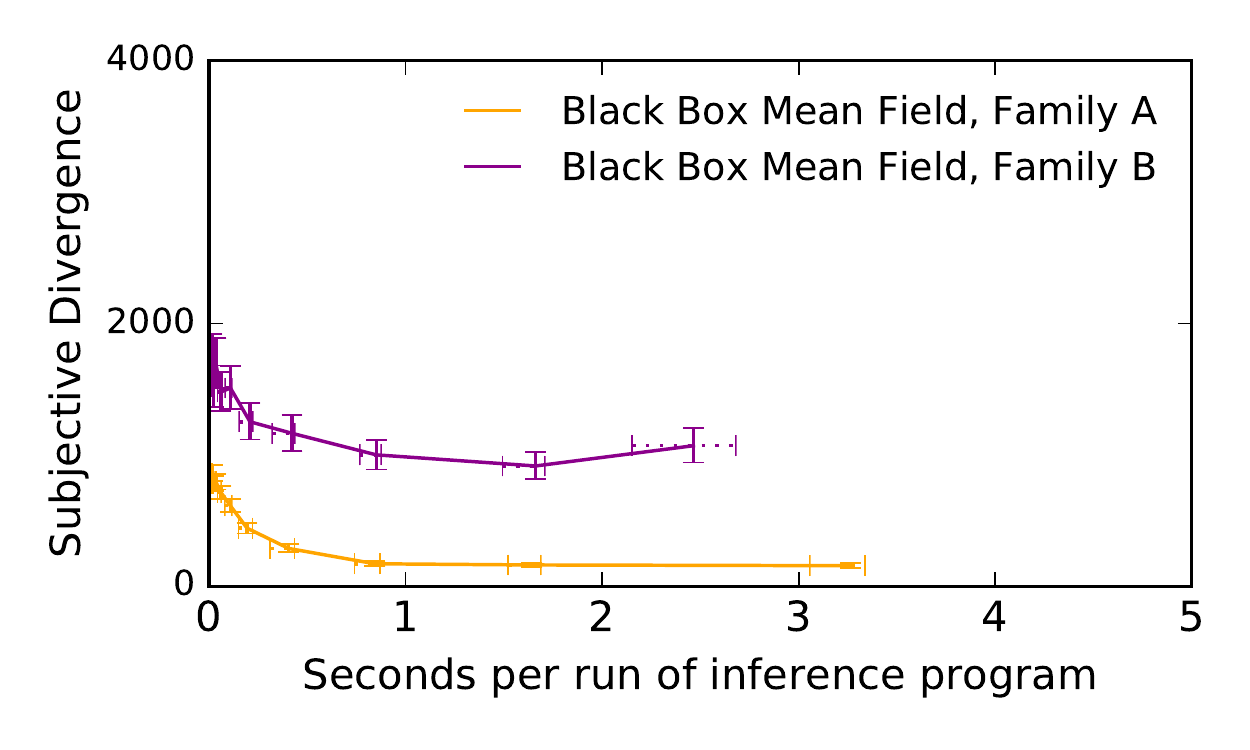}
\vspace{-7mm}
        \caption{}
        \label{fig:linreg_optimization}
    \end{subfigure}

\vspace{3mm}
\begin{minipage}{0.45\textwidth}

\begin{subfigure}[b]{1.0\textwidth}
\begin{lstlisting}[frame=tlrb,moredelim={[is][\bfseries]{@@}{@@}}]
model = make_model(
 prior: do(
 assume @@ x_coordinates = linspace(-5, 5, 11); @@
 assume @@ a ~ normal(0, 2.0); @@
 assume @@ b ~ normal(0, 2.0); @@
 assume @@ f = (x_coord) -> { a * x_coord + b }; @@
 ),
 observation_model: map(
 (i) -> [|@@normal(f(x_coordinates[@@$i@@]), 0.3)@@|],
 arange(11)))
\end{lstlisting}
\caption{Model program (model expressions bold)}
\label{fig:code_model}
\end{subfigure}
\begin{subfigure}[b]{1.0\textwidth}
\begin{lstlisting}[frame=tlrb]
(data: x, model_trace: z) -> {
  model_traces = map(
    (i) -> sample_trace_from_prior(model),
    arange(num_particles));
  k = uniform_discrete(num_particles)
  // replace one of the traces with z
  model_traces[k] = z; 
  y = (model_traces, k);
  return y
};
\end{lstlisting}
\caption{LW-SIR meta-inference program}
\label{fig:code_reverse_log_weight}
\end{subfigure}
\vspace{3mm}
\end{minipage}\hfill
\begin{minipage}{0.5\textwidth}
\begin{subfigure}[b]{1.0\textwidth}
\begin{lstlisting}[frame=tlrb]
(data: x) -> {
  model_traces = map(
    (i) -> sample_trace_from_prior(model),
    arange(num_particles));
  likelihoods = map(
    (model_trace) -> likelihood(model_trace, x)
    model_traces);
  k = categorical(normalize(likelihoods));
  y = (model_traces, k);
  z = model_traces[k];
  return (y, z)
};
\end{lstlisting}
\caption{LW-SIR inference program}
\label{fig:code_measurement}
\end{subfigure}
\begin{subfigure}[b]{1.0\textwidth}
\begin{lstlisting}[frame=tlrb]
(data: x, inf_trace: y, model_trace: z) -> {
 (model_traces, k) = y;
 likelihoods = map(
  (tr) -> likelihood(tr, x)
  model_traces);
 return mean(likelihoods)
};
\end{lstlisting}
\caption{LW-SIR weight estimate computation}
\label{fig:code_forward_log_weight}
\end{subfigure}
\vspace{3mm}
\end{minipage}
\begin{subfigure}[b]{1.0\textwidth}
    \centering
    \includegraphics[width=1.0\textwidth]{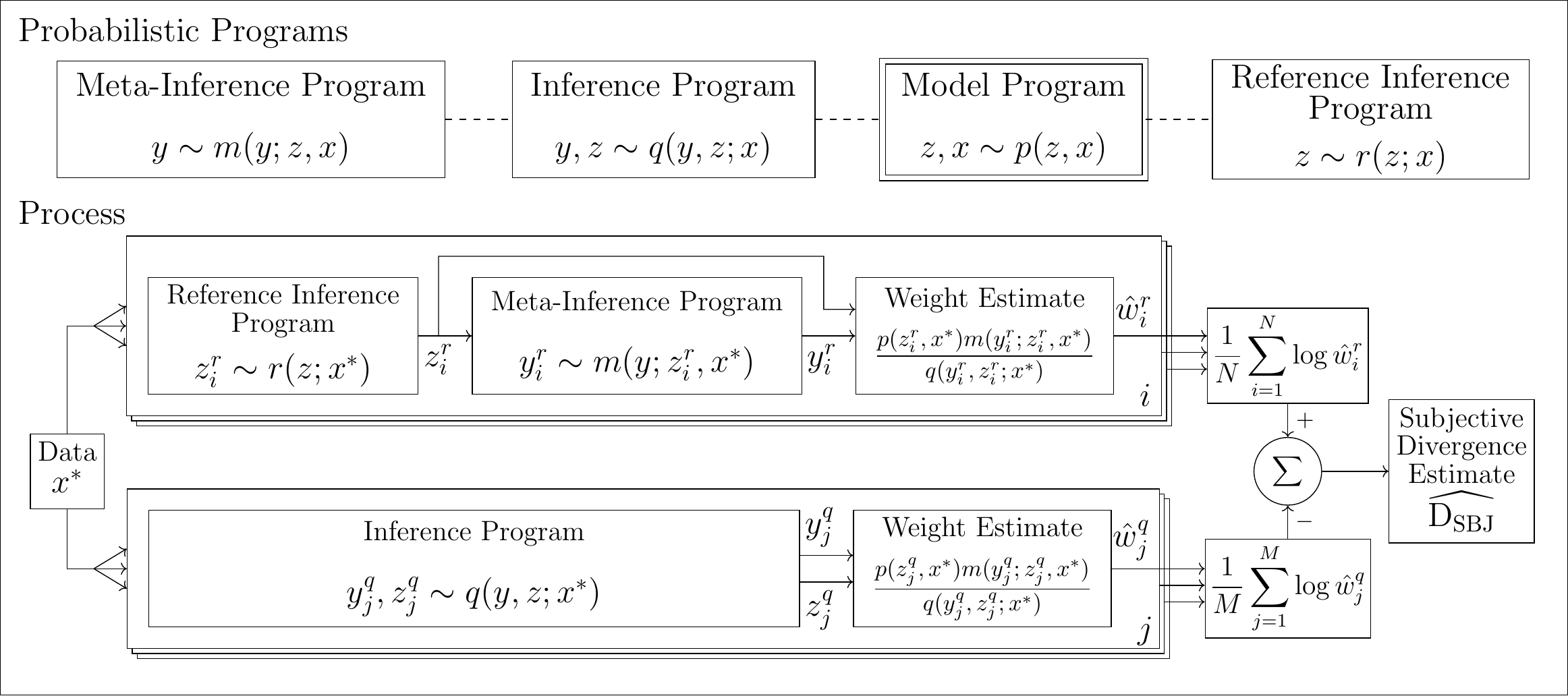}
    \caption{}
    \label{fig:schematic}
\end{subfigure}
    \caption{Estimating subjective divergences for probabilistic
inference programs. (a) relates subjective divergence and Kullback-Leibler (KL)
divergence.  (b) and (c) show estimated subjective divergence profiles for
sampling-based and optimization-based inference respectively on a small
Bayesian linear regression problem using an oracle reference program (vertical
error bars show bootstrap 90\% confidence intervals,
horizontal error bars show interquartile ranges, and $N=M=500$ runs were used for each point). (d) shows a model program
implemented in VentureScript, and (e), (f), and (g) show VentureScript programs
used to estimate subjective divergences for a sampling importance resampling
(LW-SIR) inference program. (h) shows a schematic of the probabilistic programs
and process used in Algorithm~\ref{alg:general} to estimate subjective
divergence.}
\label{fig:illustration}
\end{figure}

%% file: tex/estimating.tex
\section{Estimating subjective divergences}

Kullback-Leibler (KL) divergences between an inference program's approximating
distribution and its target distribution are objective model-independent
measures of the approximation error. However, tractable techniques for
estimating KL divergences for approximate inference are lacking.  This paper
defines a quantity, \emph{subjective divergence}, in terms of the following
elements:

\begin{enumerate}[leftmargin=*]
\item \emph{Model program} $z,x \sim p(z,x)$: Samples latent variables $z \sim p(z)$ and data $x|z \sim
p(x|z)$ for probabilistic model $p(z,x)$.
\item \emph{Data} $x^*$: A specific dataset which induces the posterior distribution
$p(z|x^*)$.
\item \emph{Approximate inference program} $y,z \sim q(y,z;x^*)$: Samples
output $z$ from $q(z;x^*)$, which approximates $p(z|x^*)$, and also returns the
history $y$ of the inference program execution that generated $z$.  An
approximate inference program induces a \emph{weight function} $w(z) :=
p(z,x^*)/q(z;x^*)$.
\item \emph{Reference inference program} $z \sim r(z;x^*)$: Gold standard
sampler that approximates the posterior $p(z|x^*)$. If the reference inference
program $r(z;x^*)$ is exact, so that $r(z;x^*) = p(z|x^*)$ for all $z$, or
equivalently $\KL(r(z;x^*)||p(z|x^*)) = 0$, we call it an \emph{oracle}.
\item \emph{Inference output marginal density estimators} $\QIS(z;x^*)$ and $\QHM(z;x^*)$:
Estimators of marginal density $q(z;x^*) = \int q(y,z;x^*) dy$ for inference
program output $z$ such that $\E \left[ \QIS(z;x^*) \right] = q(z;x^*)$ and $\E
\left[ 1/ \QHM(z;x^*) \right] = 1/q(z;x^*)$ for all $z$ and $x^*$.
$\QIS(z;x^*)$ and $\QHM(z;x^*)$ denote random variables. A realized estimate of
$q(z;x^*)$ induces a realized weight estimate.  In all cases, expectations without a
subscript are with respect to $\QIS(z;x^*)$ or $\QHM(z;x^*)$.
\end{enumerate}

\begin{frm-def}[Subjective Divergence]
~\\\resizebox{1.00\hsize}{!}{$
\SBJ(q(z;x^*)||p(z|x^*)) := \E_{z \sim r(z;x^*)} \left[ \E \left[\log \frac{p(z,x^*)}{\QIS(z;x^*)} \right] \right] - \E_{z\sim q(z;x^*)} \left[ \E \left[\log \frac{p(z,x^*)}{\QHM(z;x^*)}\right] \right]
$}
\end{frm-def}
\begin{frm-prop} If an oracle reference program is used (where $r(z;x^*) = p(z|x^*)$ for all $z$), then
~\\$\SBJ(q(z;x^*)||p(z|x^*)) \ge \KL(q(z;x^*)||p(z|x^*)) + \KL(p(z|x^*)||q(z;x^*))$
\label{prop:symmetric_bound}
\end{frm-prop} 
\sloppy This proposition is proven in Section~\ref{sec:analyzing}. To construct
inference output marginal density estimators $\QIS(z;x^*)$ and $\QHM(z;x^*)$,
we make use of a \emph{meta-inference program} $y \sim m(y;z,x^*)$, which
samples inference program execution history $y$ from an approximation to the
conditional distribution $q(y|z;x^*)$ given inference program output $z$, such
that the ratio of densities $q(y,z;x^*) / m(y;z,x^*)$ can be efficiently
computed given $y$, $z$, and $x^*$. The baseline $\QIS(z;x^*)$ estimator
samples $y \sim m(y;z,x^*)$ and produces a single sample importance sampling
estimate: $q(y,z;x^*)/m(y;z,x^*)$. The baseline $\QHM(z;x^*)$ estimator obtains
$y \sim q(y|z,x^*)$ from the history of the inference program execution that
generated $z$, and produces a single sample harmonic mean estimate:
$q(y,z;x^*)/m(y;z,x^*)$. A procedure for estimating subjective divergence using
these baseline meta-inference based estimators is shown in
Algorithm~\ref{alg:general}.

\begin{algorithm}
    \caption{Subjective divergence estimation for general inference programs}
    \label{alg:general}
    \begin{algorithmic}[1]
        \Require{Elements 1-4, meta-inference program $y \sim m(y;z,x^*)$, number of reference replicates $N$, number of inference replicates $M$}
        \For{$i \gets 1 \textrm{ to } N$} \Comment{$N$ independent replicates using reference inference program $r(z;x^*)$}
            \Sample{$z^r_i$}{$r(z;x^*)$} \Comment{Gold standard sample $z^r_i$ from reference inference program}
            \Sample{$y^r_i$}{$m(y;z^r_i,x^*)$} \Comment{Plausible inference program execution history producing $z^r_i$}
            \Let{$\hat{q}^r_i$}{$\frac{q(y^r_i,z^r_i;x^*)}{m(y^r_i;z^r_i,x^*)}$} \Comment{Estimate of marginal output density $q(z^r_i;x^*)$}
            \Let{$\hat{w}^r_i$}{$\frac{p(z^r_i,x^*)}{\hat{q}^r_i}$} \Comment{Estimate of weight}
        \EndFor
        \For{$j \gets 1 \textrm{ to } M$} \Comment{$M$ independent replicates using inference program $q(z;x^*)$}
            \Sample{$y^q_j,z^q_j$}{$q(y,z;x^*)$} \Comment{Output $z^q_j$ and execution history $y^q_j$ from inference program}
            \Let{$\hat{q}^q_j$}{$\frac{q(y^q_j,z^q_j;x^*)}{m(y^q_j;z^q_j,x^*)}$} \Comment{Estimate of marginal output density $q(z^q_j;x^*)$}
            \Let{$\hat{w}^q_j$}{$\frac{p(z^q_j,x^*)}{\hat{q}^q_j}$} \Comment{Estimate of weight}
        \EndFor
        \State \Return{$\frac{1}{N}\sum_{i=1}^N \log \hat{w}^r_i - \frac{1}{M} \sum_{j=1}^M \log \hat{w}^q_j$}
    \end{algorithmic}
\end{algorithm}

\begin{figure}[t]
    \centering
    \begin{minipage}{0.35\textwidth}
        \begin{subfigure}[b]{1.0\textwidth}
            \includegraphics[width=1.0\textwidth]{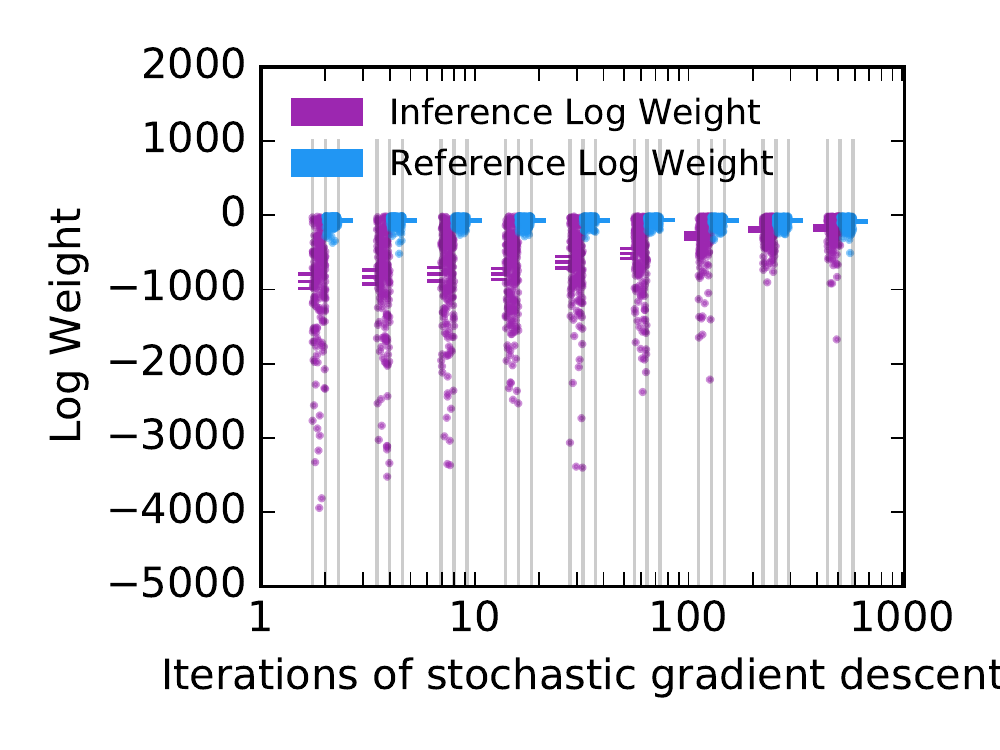}
            \caption{}
            \label{fig:raw_data}
        \end{subfigure}
        \begin{subfigure}[b]{1.0\textwidth}
            \includegraphics[width=1.0\textwidth]{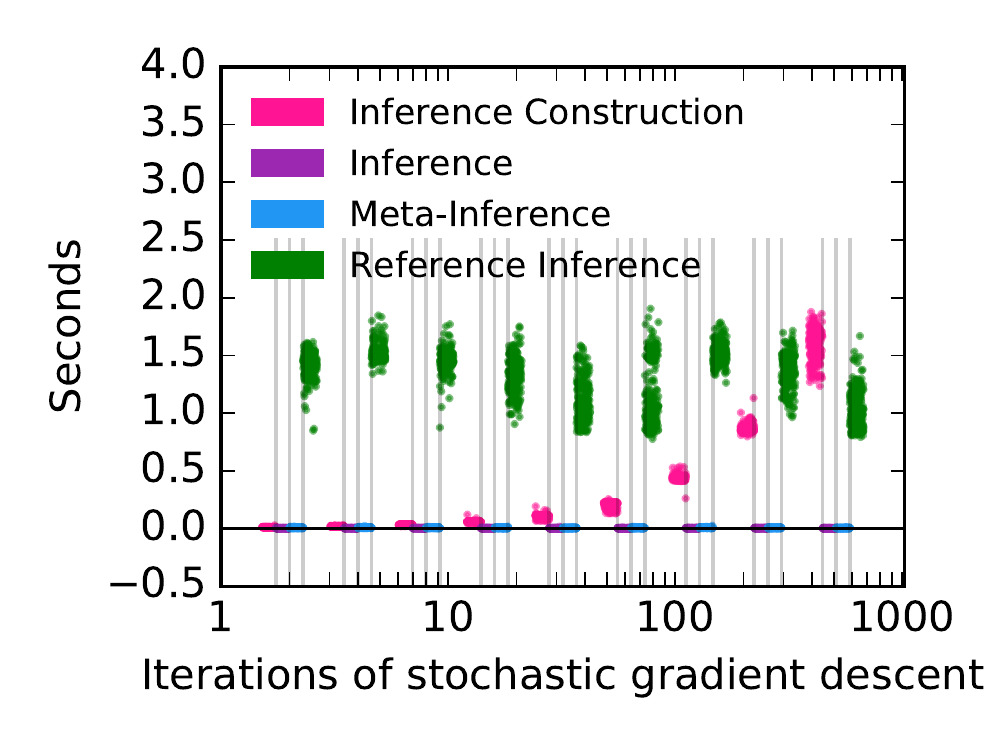}
            \caption{}
        \end{subfigure}
    \end{minipage}\hfill
    \begin{minipage}{0.6\textwidth}
    \begin{subfigure}[b]{1.0\textwidth}
\resizebox{1.0\textwidth}{!}{%
\includegraphics[width=1.0\textwidth]{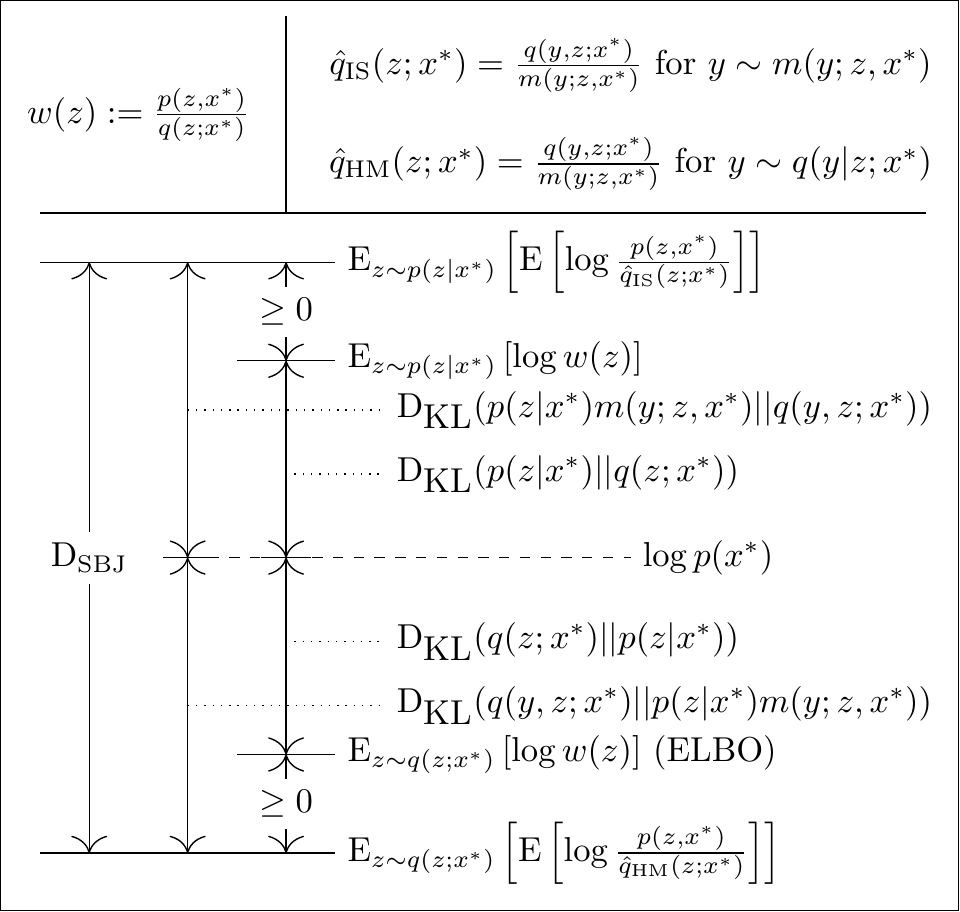}
}
        \caption{}
        \label{fig:bounds_schematic}
    \end{subfigure}
    \end{minipage}
    \caption{Raw log estimated weight data (a) and timing data (b) for
individual runs of the VentureScript implementation of
Algorithm~\ref{alg:general} on a black box variational inference program.
Weights are colored by the source of $z$ (inference or reference program), and
timing data is broken down into stages of the estimation procedure. (c) shows a
schematic illustration of the relationship between key KL divergences and
subjective divergence estimated by Algorithm~\ref{alg:general} in the case when
the reference program is an exact inference oracle.}
\end{figure}

We have produced a VentureScript inference programming library that implements
Algorithm~\ref{alg:general}. In our applications, the weight estimate computation can be
performed incrementally within the inference program and meta-inference
program, obviating the need for an explicit representation of inference program
execution history and the separate weight computation illustrated in
Figure~\ref{fig:illustration}. 

The subjective divergence is based on estimating the \emph{symmetrized} KL
divergence in order to handle the fact that the posterior density is only
available in unnormalized form (the
symmetrized KL can be expressed purely in terms of unnormalized densities). We
use a reference sampler as a proxy for a posterior sampler (accepting
subjectivity) to address the challenge of Monte Carlo estimation with respect
to the posterior $p(z|x^*)$ for the term $\KL(p(z|x^*)||q(z;x^*))$ in the
symmetrized KL. For inference programs with an output density $q(z;x^*)$ that
can be computed efficiently, such as mean-field variational families, the
weight estimate in Algorithm~\ref{alg:general} can be replaced with the 
true weight $p(z,x^*) / q(z;x^*)$, and the subjective divergence is
equivalent to the symmetrized KL divergence when an oracle reference is used.
For inference programs with a large number of internal random
choices $y$, the densities on outputs $q(z;x^*)$ are intractable to compute,
and Algorithm~\ref{alg:general} uses meta-inference to construct marginal
density estimators $\QIS(z;x^*)$ and $\QHM(z;x^*)$ such that
Proposition~\ref{prop:symmetric_bound} holds. Subjective divergence can be
interpreted as approximately comparing samples from the inference program of interest to
gold standard samples through the lens of the log-weight function $\log
w(\cdot)$.

%% file: tex/analyzing.tex
\section{Analyzing subjective divergences}
\label{sec:analyzing}

Having defined subjective divergence and a procedure for estimating it, we now
prove Prop.~\ref{prop:symmetric_bound} using bounds on the expected log estimated weight taken under the inference program of interest
($\E_{z\sim q(z;x^*)} \left[ \E \left[\log (p(z,x^*)/\QHM(z;x^*))\right] \right]$) and the expected log estimated weight taken under the reference program ($\E_{z \sim r(z;x^*)} \left[ \E \left[\log (p(z,x^*)/\QIS(z;x^*)) \right] \right]$).
The expectation under the inference program of interest is less than the log normalizing constant $\log p(x^*)$
by at least $\KL(q(z;x^*)||p(z|x^*))$:
\begin{frm-lemma}
$\E_{z\sim q(z;x^*)} \left[ \E \left[ \log \frac{p(z,x^*)}{\QHM(z;x^*)} \right] \right] \le \log p(x^*) - \KL(q(z;x^*)||p(z|x^*))$
\label{lemma:inference_procedure_bound}
\end{frm-lemma}
\vspace{-2mm}
(derivation based on Jensen's inequality in Appendix~\ref{sec:proofs}).
Note that this constitutes a lower bound on the ``ELBO'' variational objective.
The expectation under an oracle reference program is
greater than the log normalizing constant by at least
$\KL(p(z|x^*)||q(z;x^*))$:
\begin{frm-lemma}
$\E_{z\sim p(z|x^*)} \left[ \E \left[ \log \frac{p(z,x^*)}{\QIS(z;x^*)} \right] \right] \ge \log p(x^*) + \KL(p(z|x^*)||q(z;x^*))$
\label{lemma:reference_procedure_bound}
\end{frm-lemma}
\vspace{-3mm}
(derivation based on Jensen's inequality in Appendix~\ref{sec:proofs}).
The subjective divergence is the difference between the expectation under the
reference program (bounded in Lemma~\ref{lemma:reference_procedure_bound}) and the expectation under the inference program of interest (bounded in Lemma~\ref{lemma:inference_procedure_bound}).
Taking the difference of the bounds cancels the $\log p(x^*)$ terms and proves Proposition~\ref{prop:symmetric_bound}.
Relationships between key quantities in the proof are illustrated in Figure~\ref{fig:bounds_schematic}.
By Proposition~\ref{prop:symmetric_bound}, if an oracle reference program is
available, we can estimate an \emph{upper bound} on the symmetrized KL
divergence by estimating a subjective divergence.
\begin{figure}[t]
    \centering
    \begin{minipage}{0.33\textwidth}
        \begin{subfigure}[b]{1.0\textwidth}
            \centering
            \includegraphics[width=1.0\textwidth]{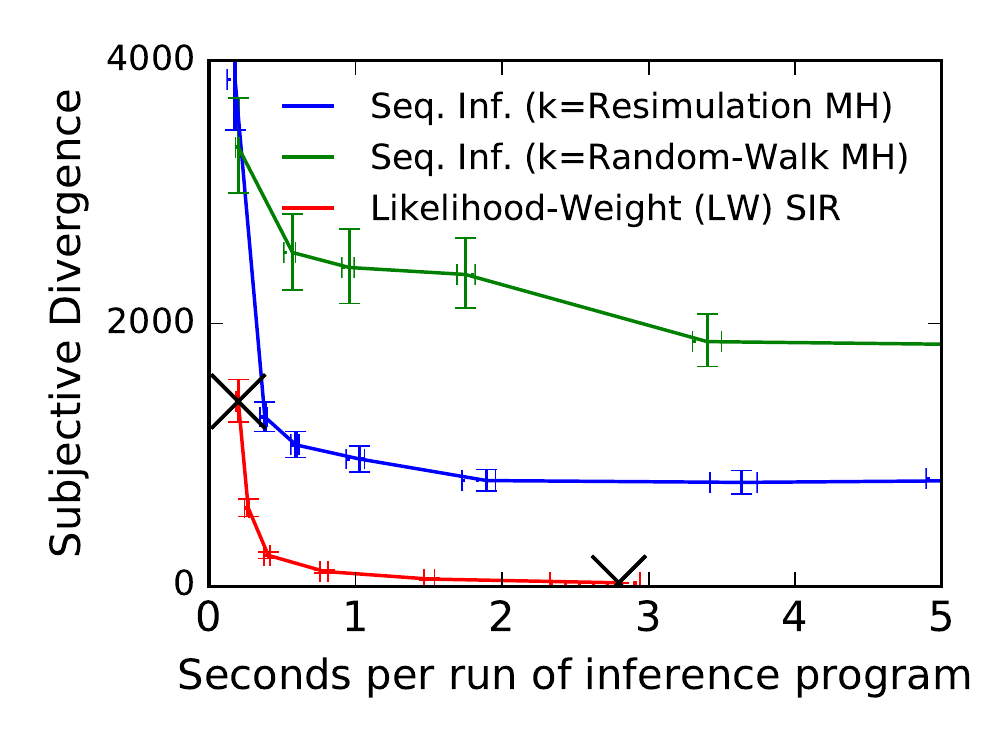}
            \caption{Oracle Reference}
        \end{subfigure}
    \end{minipage}\hfill
    \begin{minipage}{0.33\textwidth}
        \begin{subfigure}[b]{1.0\textwidth}
            \centering
            \includegraphics[width=1.0\textwidth]{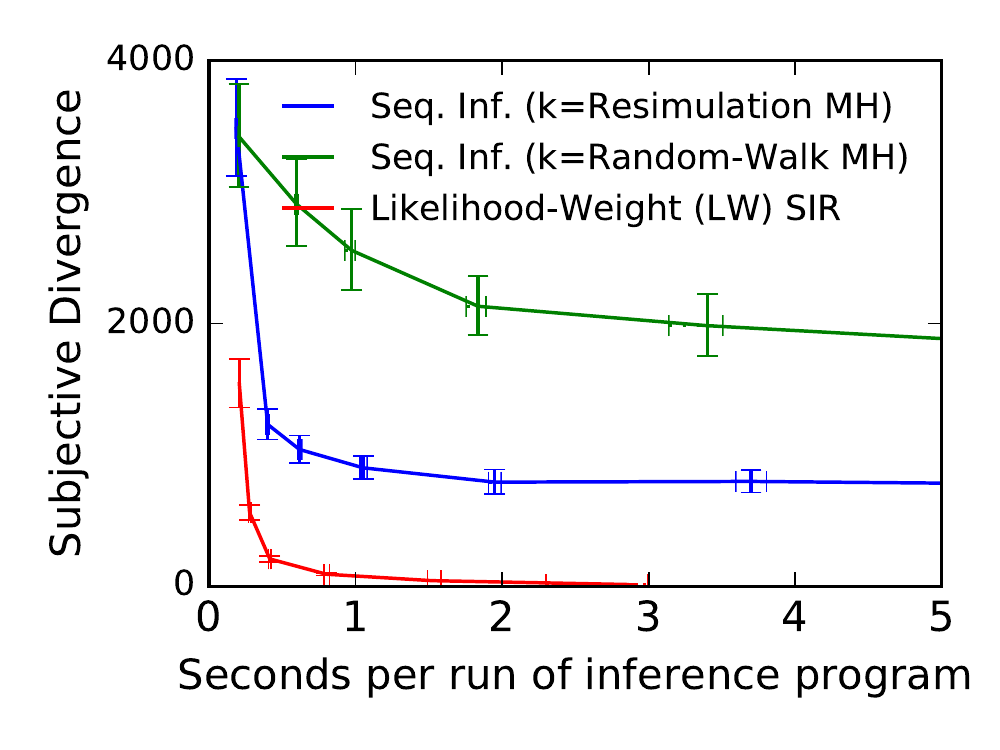}
            \caption{LW-SIR (64) Reference}
        \end{subfigure}
    \end{minipage}\hfill
    \begin{minipage}{0.33\textwidth}
        \begin{subfigure}[b]{1.0\textwidth}
            \centering
            \includegraphics[width=1.0\textwidth]{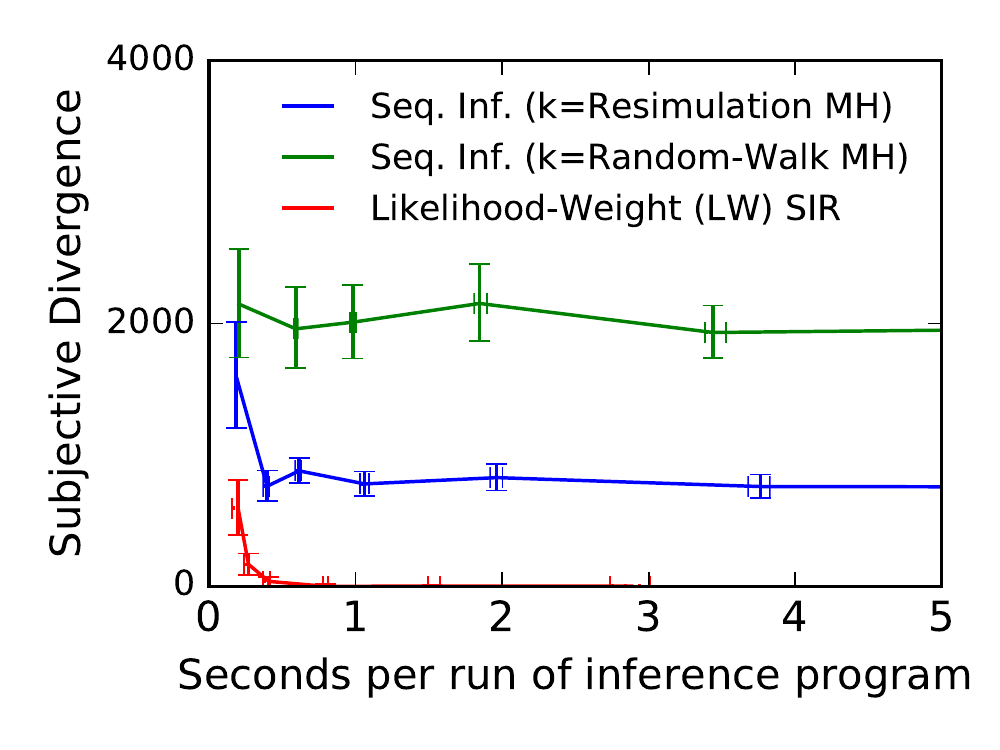}
            \caption{LW-SIR (2) Reference}
        \end{subfigure}
    \end{minipage}
    \caption{Subjective divergence profiles for several inference programs on
the Bayesian linear regression problem of Figure~\ref{fig:illustration}, using
an oracle reference (a), a high quality approximate reference (b) and a low
quality approximate reference (c). The oracle implements collapsed sampling
from the posterior. The LW-SIR (sampling importance resampling with prior
importance distribution) references use 64 and 2 particles
respectively, and their own subjective divergence estimates using the oracle
reference are marked with the symbol $\times$ in (a). The profiles based on the oracle reference and the
high quality LW-SIR (64) reference are qualitatively similar.}
    \label{fig:compare_oracles_linreg}
\end{figure}

\subsection{Effect of quality of reference inference program}
\vspace{-1mm}
If the reference inference program $r(z;x^*)$ is not an oracle, it is
still possible to retain the validity of subjective divergence as an upper
bound, depending on the accuracy of the reference program:
\begin{frm-prop} If $\KL(r(z;x^*)||p(z|x^*)) \le \KL(r(z;x^*)||q(z;x^*)) - \KL(p(z|x^*)||q(z;x^*))$
then $\SBJ(q(z;x^*)||p(z|x^*)) \ge \KL(q(z;x^*)||p(z|x^*)) + \KL(p(z|x^*)||q(z;x^*))$
\end{frm-prop}
\vspace{-2mm}
\begin{frm-prop} If $\KL(r(z;x^*)||p(z|x^*)) \le \KL(r(z;x^*)||q(z;x^*))$ then
\begin{align*}
\E_{z\sim r(z;x^*)} \left[ \E \left[ \log \frac{p(z,x^*)}{\QIS(z;x^*)} \right] \right] &\ge \log p(x^*)\\
\SBJ(q(z;x^*)||p(z|x^*)) &\ge \KL(q(z;x^*)||p(z|x^*))
\end{align*}
\end{frm-prop}
\vspace{-1mm}
(derivations in Appendix~\ref{sec:proofs}). When the density $q(z;x^*)$ is available, we use $\QIS(z;x^*) = \QHM(z;x^*) = q(z;x^*)$, and
$\SBJ$ replaces the $\KL(p(z|x^*)||q(z;x^*))$ term in the symmetrized KL
divergence with $\KL(r(z;x^*)||q(z;x^*)) -
\KL(r(z;x^*)||p(z|x^*))$.
Figure~\ref{fig:compare_oracles_linreg} compares subjective divergence
profiles obtained using oracle reference and approximate inference
references of varying quality. 

\subsection{Effect of quality of meta-inference program}
\vspace{-1mm}
In the setting of an oracle reference program and the baseline marginal
density estimators $\QIS(z;x^*)$ and $\QHM(z;x^*)$ used in Algorithm~\ref{alg:general}, the
quality of the meta-inference $m(y;z,x^*)$ determines the tightness of the
upper bound of Proposition~\ref{prop:symmetric_bound}.  In particular, the gap
between the true symmetrized KL divergence and the subjective divergence is the
symmetrized conditional relative entropy \cite{cover2012elements}
(derivation in Appendix~\ref{sec:metainference_analysis})
\vspace{-1mm}
\begin{equation} \label{eq:general_metainference_gap}
\E_{z\sim q(z;x^*)} \left[ \KL(q(y|z;x^*)||m(y;z,x^*)) \right] + \E_{z\sim p(z|x^*)} \left[ \KL(m(y;z,x^*)||q(y|z;x^*)) \right]
\end{equation}
which measures how closely the meta-inference approximates $q(y|z;x^*)$, the
conditional distribution on inference execution histories given inference output $z$.
Note that if we had exact meta-inference \emph{and} could compute its density,
the weight estimate simplifies to $w(z) := p(z,x^*) / q(z;x^*)$, and we could
remove this gap. More generally, the gap is due to the biases of the estimators
for $\log q(z;x^*)$ that are induced by taking the $\log(\cdot)$ of the
estimates of $q(z;x^*)$ produced by $\QIS(z;x^*)$ and $\QHM(z;x^*)$, which are related to
the variances of $\QIS(z;x^*)$ and $1/\QHM(z;x^*)$.  For example,
compare the variance of the baseline $\QIS(z;x^*)$ with the bias of the induced
estimator of $\log q(z;x^*)$:
\vspace{-2mm}
\begin{align}
\Var\left( \frac{\QIS(z;x^*)}{q(z;x^*)} \right) &= \E \left[\left(\frac{\QIS(z;x^*)}{q(z;x^*)}\right)^2 - 1 \right] = \chi_P^2 (m(y;z,x^*)||q(y|z;x^*))\\
\log q(z;x^*) - \E \left[ \log \QIS(z;x^*) \right] &= \E \left[ \log \frac{q(z;x^*)}{\QIS(z;x^*)} \right] = \KL(m(y;z,x^*)||q(y|z;x^*))
\end{align}
where $\chi_P^2(m(y;z,x^*)||q(y|z;x^*))$ is the Pearson chi-square divergence
\cite{nielsen2013chi}. The bias of the estimator of $\log q(z;x^*)$ manifests
in the second term in Equation~\ref{eq:general_metainference_gap}.
\hspace{-3mm}
\footnote{Improving upon the baseline inference output marginal density estimators and reducing
the gap between subjective divergence and symmetric KL divergence seems a
promising direction for future work.}
See Appendix~\ref{sec:metainference_analysis} for details.

\subsection{Related work}
\label{sec:related}
In \cite{grosse2015sandwiching} the authors point out that unbiased estimators
like $\QIS(z;x^*)$ and unbiased reciprocal estimators like $\QHM(z;x^*)$
estimate lower and upper bounds of the log-estimand respectively, which they
use to estimate lower and upper bounds on $\log p(x^*)$.
\cite{grosse2015sandwiching} also suggests combining stochastic upper bounds on
$\log p(x^*)$, obtained by running reversed versions of sequential Monte Carlo
(SMC) algorithms starting with an exact sample obtained when simulating data
$x^*$ from the model, with lower bounds on the ELBO, to upper bound KL
divergences.  The authors of \cite{DBLP:conf/icml/SalimansKW15} introduce a general
auxiliary variable formalism for estimating lower bounds on the
ELBO of Markov chain inference, which is equivalent to estimation of our expected log estimated weight under the inference program for the baseline $\QHM$ estimator applied to Markov chains.

%% file: tex/applications.tex
\vspace{-2mm}
\section{Applications}
\label{sec:applications}

We used the VentureScript implementation of Algorithm~\ref{alg:general} to
estimate subjective divergence profiles for diverse approximate inference
programs applied to several probabilistic models.

In addition to applying the technique to mean-field variational inference, where
the output density $q(z;x^*)$ is available, we derived
meta-inference programs for two classes of inference programs whose density
is generally intractable: sequential inference utilizing a Markov chain
of detailed-balance transition operators and particle filtering in state space
models. For sequential inference, we use a coarse-grained representation of the
inference execution history that suppresses internal random choices made within segments
of the Markov chain that satisfy detailed balance with respect to a single
distribution. The meta-inference program is also sequential detailed-balance
inference, but with the order of the transition operators reversed. This
reversed Markov chain is an instance of the formalism of
\cite{DBLP:conf/icml/SalimansKW15}, was used to construct annealed importance
sampling (AIS) \cite{Neal2001}, and was sampled from in
\cite{burda2014accurate} and \cite{grosse2015sandwiching}.  The weight estimate
corresponds to the AIS marginal likelihood
estimate. The subjective divergence for standard non-sequential MCMC can be
analyzed using this construction, but results in a trivial upper bound on the
KL divergence due to the failure of the approximating assumptions used to
derive the meta-inference program.  For particle filtering in state space
models, we use the conditional SMC (CSMC) update \cite{andrieu2010particle} and
the weight estimate is the marginal likelihood estimate of the particle filter.
It is intuitive that we use CSMC to answer ``how might have a particle filter
produced a given particle?'' A special case of the particle filter is sampling
importance resampling, for which the meta-inference
program (shown in Figure~\ref{fig:code_reverse_log_weight}) places the output sample $z$ in one of $K$ particles, and samples the
remaining $K-1$ particles from the prior. See Appendix~\ref{sec:app_derivations} for derivations.

\subsection{Linear regression}
We first considered a small Bayesian linear regression problem, with unknown
intercept and slope latent variables (model program shown in
Figure~\ref{fig:code_model}), and generated subjective divergence profiles for sampling-based and variational
inference programs (shown in Figure~\ref{fig:linreg_sampling} and Figure~\ref{fig:linreg_optimization})
using an oracle reference. 
We estimated profiles for two black box mean-field \cite{DBLP:conf/aistats/RanganathGB14} programs which
differed in their choice of variational family---each family had a different
fixed variance for the latents. We varied the number of iterations of
stochastic gradient descent to generate the profiles, which exhibited distinct
nonzero asymptotes. 
We also estimated profiles for two sequential inference programs that consist
of alternating between observing an additional data point $x^*_t$ and running a
transition operator $k_t$ that targets the partial posterior $p(z | x^*_{1:t})$
for $t=1\ldots, T$ with $T = 11$ data points. One program used
repeated application of Metropolis-Hastings (MH) transitions with a
resimulation (prior) proposal within each $k_t$ and the other used 
applications of a random-walk MH transition. We varied the number of
applications within each $k_t$ of the primitive MH transition operator. The
profile based on resimulation MH converged more rapidly. 
Finally, we produced a subjective divergence profile for a likelihood-weighting sampling
importance resampling (LW-SIR) inference program by varying the number of
particles.  LW-SIR was the only algorithm applied to this problem whose
subjective divergence profile converged to zero.
\hspace{-3mm}
\footnote{The profiles for the sequential detailed balance inference scheme converge to the sum of symmetrized KL divergences between consecutive partial
posteriors $p(z), p(z|x^*_1), p(z|x^*_{1:2}), \ldots, p(z|x^*_{1:T})$. See Appendix~\ref{sec:sequential_inference} for details.}
\begin{figure}[t]
    \centering
    \begin{minipage}{0.15\textwidth}
         \begin{subfigure}[b]{1.0\textwidth}
            \centering
            \includegraphics[width=1.0\textwidth]{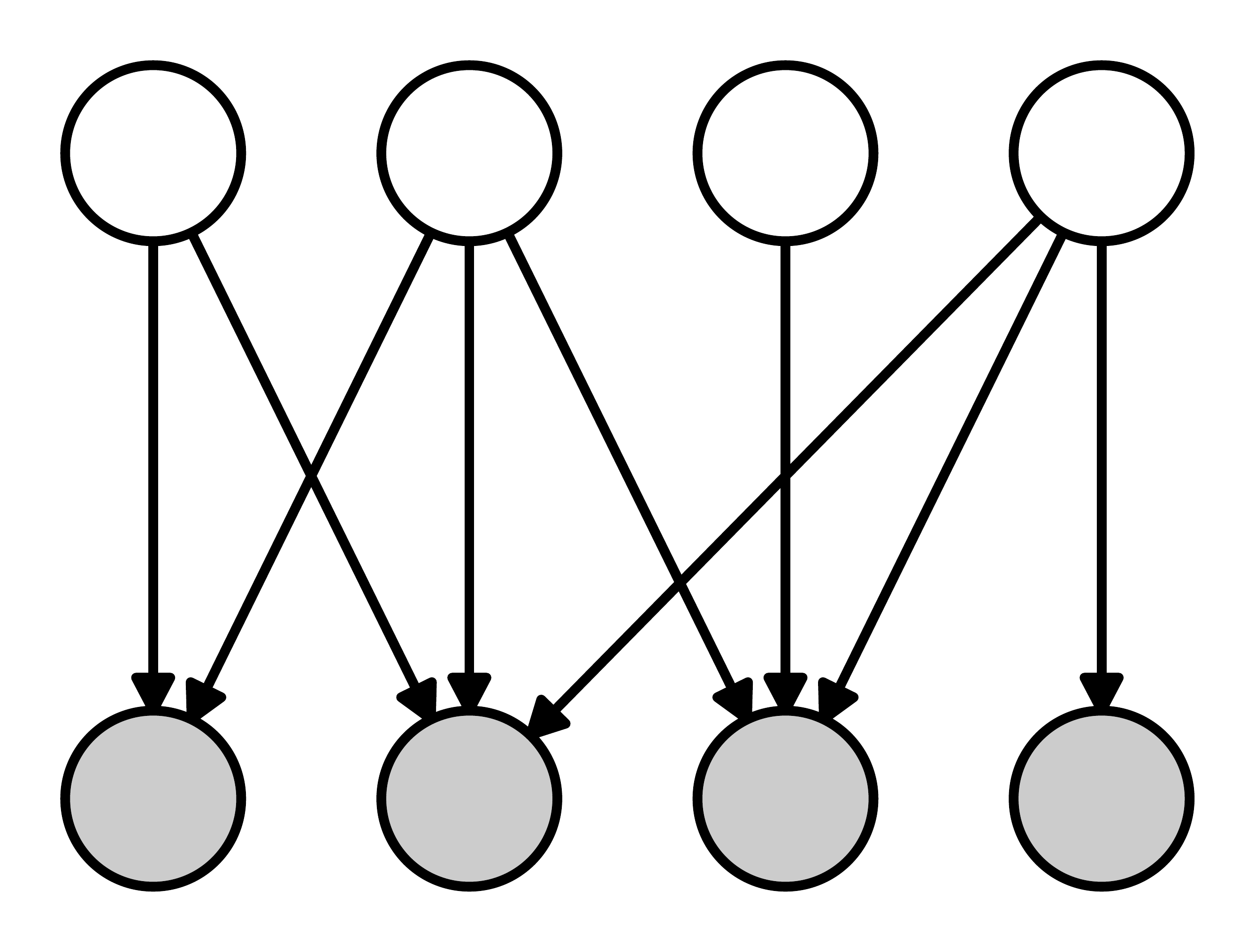}
            \caption{}
            \label{fig:qmr_network}
        \end{subfigure}
         \begin{subfigure}[b]{1.0\textwidth}
            \centering
            \includegraphics[width=1.0\textwidth]{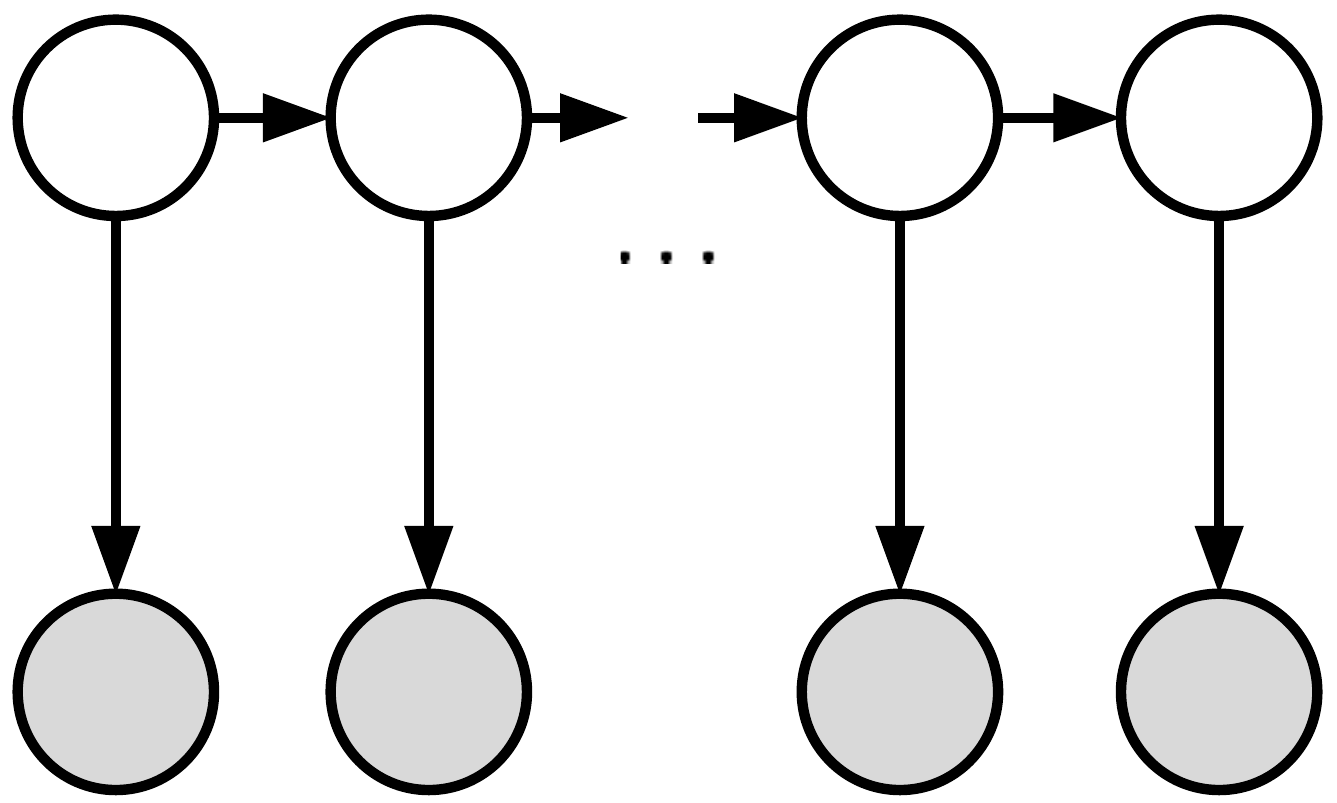}
            \caption{}
        \end{subfigure}
    \end{minipage}
    \begin{minipage}{0.4\textwidth}
        \begin{subfigure}[b]{1.0\textwidth}
            \includegraphics[width=1.0\textwidth]{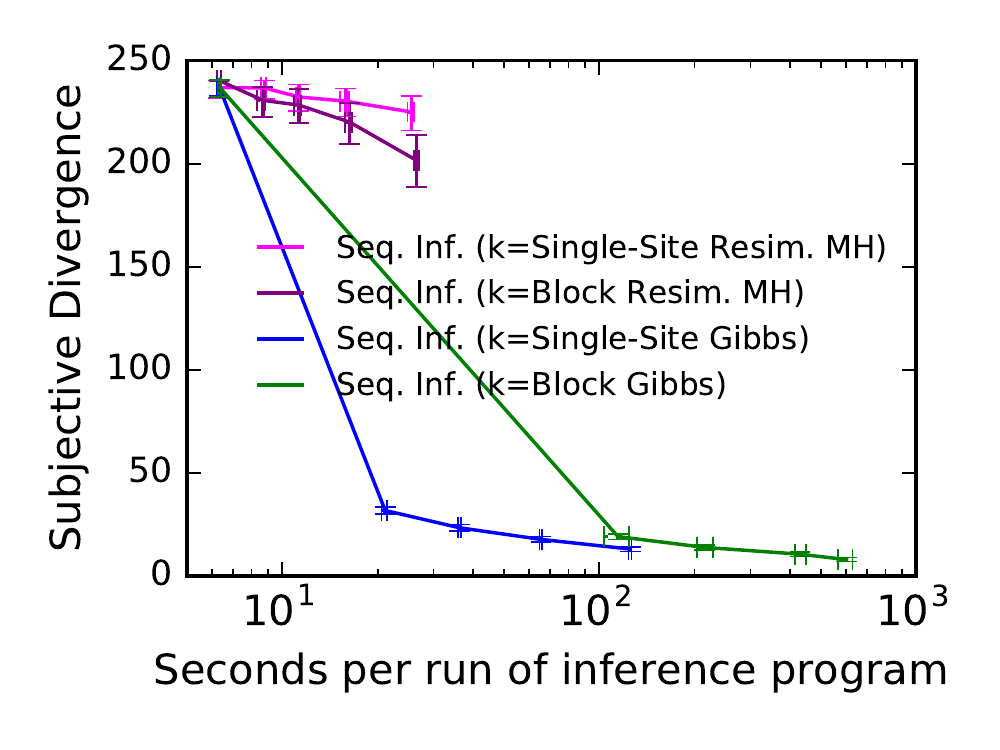}
            \caption{}
            \label{fig:qmr_plot}
        \end{subfigure}
    \end{minipage}
    \begin{minipage}{0.4\textwidth}
        \begin{subfigure}[b]{1.0\textwidth}
            \centering
            \includegraphics[width=1.0\textwidth]{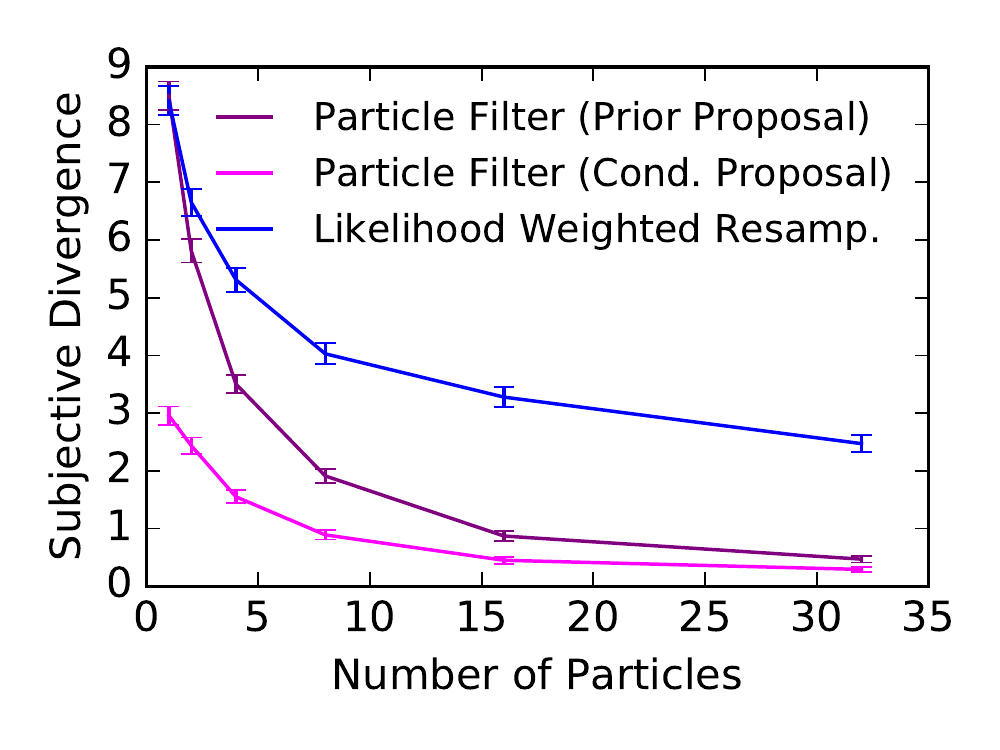}
            \caption{}
            \label{fig:particle_filter_plot}
        \end{subfigure}
    \end{minipage}
    \caption{(a) shows a subset of a noisy-or network, and (c) shows
subjective divergence profiles of sequential inference programs based on
single-site and block transition operators in the network, using a block
Gibbs-based scheme as the reference. (b) shows a subset of an HMM and (d) shows subjective divergence profiles for two particle
filters with different proposals and (non-sequential) likelihood-weighted resampling applied to the smoothing problem in the HMM
using an exact oracle reference.}
    \label{fig:compare_oracles_qmr}
\end{figure}
\vspace{-2mm}
\subsection{Bayesian networks}
\vspace{-2mm}
We estimated subjective divergence profiles for approximate inference programs
applied to a noisy-or Bayesian network (subset shown in
Figure~\ref{fig:qmr_network}). The network contained 25 latent causes, and 35
findings, with prior cause probabilities of 0.001, transmission probabilities
of 0.9, and spontaneous finding activation probabilities of 0.001, with edges
sampled uniformly with probability 0.7 of presence. All findings were active.
We compared four sequential inference programs that all
advanced through the same sequence of target distributions defined by gradually
lowering the finding spontaneous activation probability from 0.99 to the true
model value 0.001 across 10 equal-length steps, but applied distinct types of
transition operators $k_t$ at each step. We compared the use of a single-site
resimulation MH operator, a block resimulation MH operator, single-site Gibbs
operator, and block Gibbs operator as primitive operators within each $k_t$ for
$t=1,\ldots,10$, and varied the number of applications of each primitive
operator within each $k_t$ to generate the profiles,
shown in Figure~\ref{fig:qmr_plot}. For the reference program we used
sequential inference with four applications of block Gibbs between each target
distribution step. Inference for this problem is hard for single-site Gibbs
operators due to explaining away effects, and hard for resimulation-based
operators due to the low probability of the data under the prior. The
resimulation MH based profiles exhibited much slower convergence than those of
the Gibbs operators.

\vspace{-1mm}
\subsection{Hidden Markov models}
\vspace{-2mm}
We next applied the technique to a hidden Markov model (HMM) with discrete state and
observation space (40 time steps, 2 hidden states, 3 observation states), and
produced subjective divergence profiles for two particle filter inference
programs with prior (forward simulation) and conditional proposals. Both
particle filters used independent resampling. We used exact forward-filtering
backwards sampling for the reference inference program.  The profiles
with respect to the number of particles are shown in
Figure~\ref{fig:particle_filter_plot}. The conditional proposal profile
exhibits faster convergence as expected. Note that for the single particle case
there are no \emph{latent} random choices $y$ in these the particle filters, and the
subjective divergence is the symmetrized KL divergence.

\vspace{-1mm}
\subsection{Detecting an ergodicity violation in samplers for Dirichlet process mixture modeling}
\begin{figure}[t]
\centering
\begin{subfigure}[b]{0.45\textwidth}
    \centering
    \includegraphics[width=1.0\textwidth]{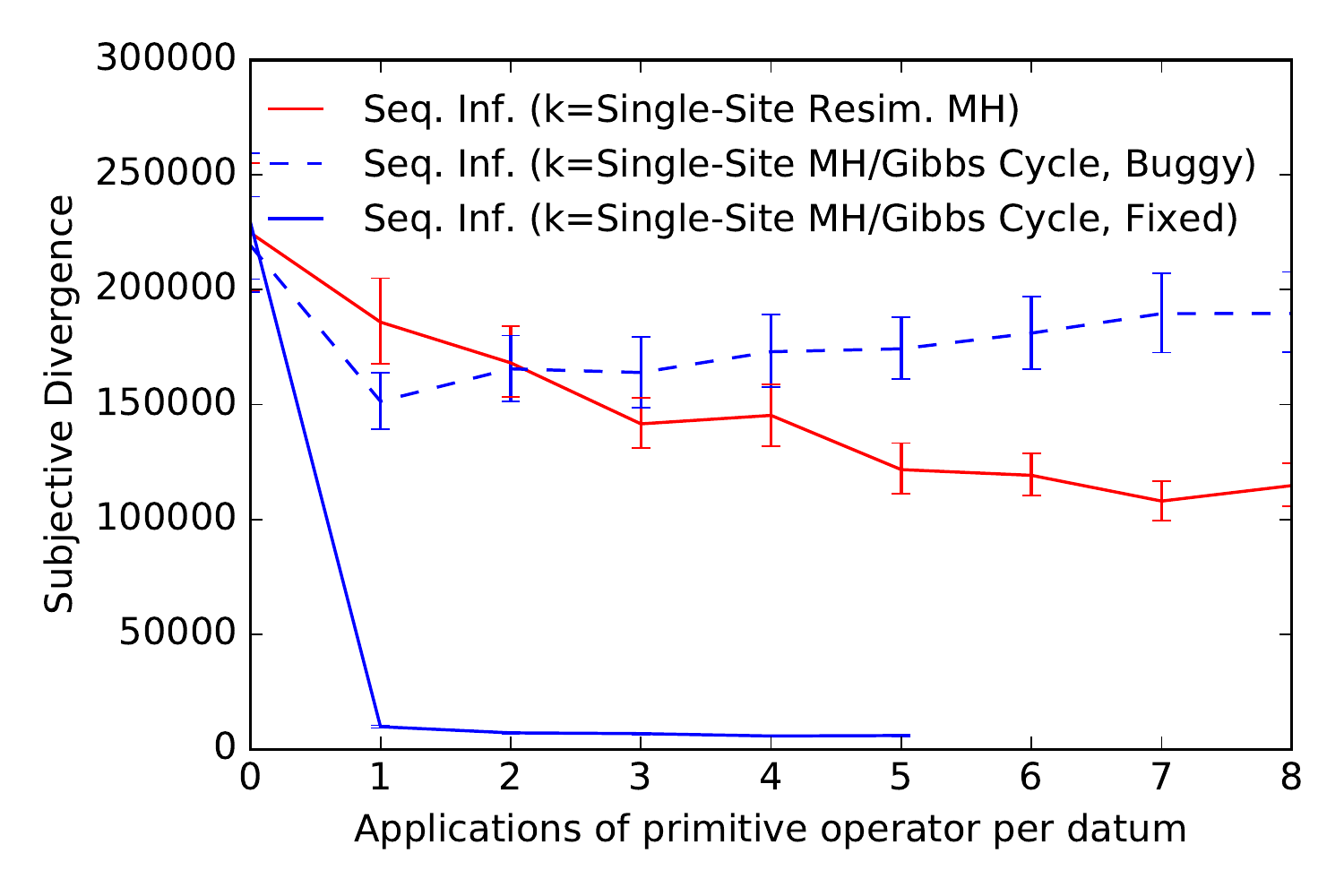}
    \caption{Subjective divergence profiles (apparent bug: subjective divergence not improving)}
    \label{fig:dpmm_bug_divergence}
\end{subfigure}%
\hspace{5mm}
\begin{subfigure}[b]{0.45\textwidth}
    \centering
    \includegraphics[width=1.0\textwidth]{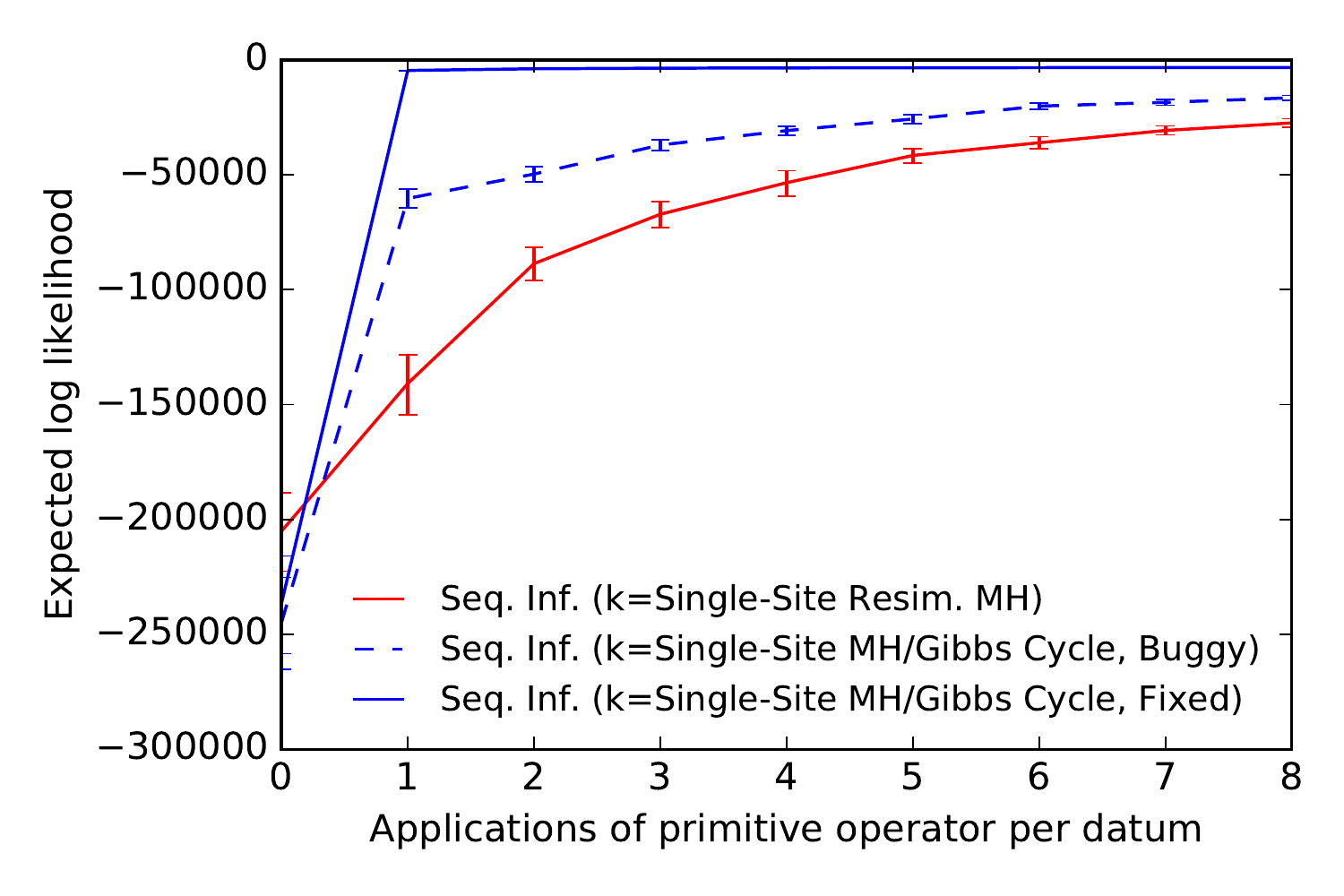}
    \caption{Expected log likelihood profiles (no apparent bug: Gibbs/MH outperforms vanilla MH)}
    \label{fig:dpmm_bug_log_likelihood}
\end{subfigure}%
    \caption{Comparing the effect of a bug in a MH transition operator
implementation on subjective divergence profiles and on profiles of the
expected log likelihood.}
    \label{fig:dpmm_debug}
\end{figure}

\vspace{-2mm}
We estimated subjective divergence profiles
(Figure~\ref{fig:dpmm_bug_divergence}) for sequential inference programs in an
uncollapsed Dirichlet process mixture model (DPMM) with $T =1000$ data points 
simulated from the model program, with partial posteriors $p(z|x^*_{1:t})$ for
$t=1,\ldots,T$ for the sequence of target distributions.  For the reference, we
used a relatively trusted sequential inference program
based on Venture's built-in single-site resimulation MH
implementation.  We estimated subjective divergence profiles for inference
based on the single-site resimulation MH operator and for inference
based on a cycle operator consisting of single-site Gibbs steps for
the latent cluster assignments, and resimulation MH for global parameters. The
subjective divergence of the Gibbs/MH operator exhibited anomalous behavior, and
degraded with additional inference, quickly becoming worse than the
resimulation MH operator.  This led us to identify a bug in our Gibbs/MH
operator in which no inference was being performed on the within-cluster
variance parameter. The profile for the corrected operator exhibited markedly
faster convergence than the resimulation MH profile. For comparison, we
estimated the expected log likelihood $\E_{z \sim q(z;x^*)} \left[ \log
p(x^*_{1:T}|z)\right]$ for output samples $z$ produced at the termination of these
inference programs. The expected log likelihood profile
(Figure~\ref{fig:dpmm_bug_log_likelihood}) for the Gibbs/MH operator with
a bug was significantly higher (better) than the profile for resimulation MH,
despite being significantly poorer than the profile in the corrected version.
Note that unlike the subjective divergence profiles, the expected log likelihood profiles for the operator with a bug may
not have seemed anomalous.

%% file: tex/discussion.tex
\section{Discussion}
\vspace{-2mm}

This paper introduced a new technique for quantifying the approximation error of a broad class of probabilistic inference programs. The key ideas are (i) to assess error relative to subjective beliefs in the quality of a reference inference program, (ii) to use symmetrized divergences, and (iii) to use a meta-inference program that finds probable executions of the original inference program if its output density cannot be directly assessed. The approach is implemented as a probabilistic meta-program in VentureScript that uses ancillary probabilistic meta-programs for the reference and meta-inference schemes.

Much more empirical and theoretical development is needed. Specific directions include better characterizing the impact of reference and meta-inference quality and identifying the contexts in which the theoretical bounds are predictably tight or loose. Applying the technique to a broad corpus of VentureScript programs seems like a useful first step. Empirically studying the behavior of subjective divergence for a broader sample of buggy inference programs also will be informative. 

It also will be important to connect the approach to results from theoretical computer science, including the computability \cite{ackerman2010computability} and complexity \cite{freer2010probabilistic} of probabilistic inference. For example, the asymptotic scaling of probabilistic program runtime can be analyzed using the standard random access memory model \cite{thomas2001introduction} under suitable assumptions about the implementation. This includes the model program; the inference program; the reference program; the meta-inference program; and the probabilistic meta-program implementing Algorithm 1. It should thus be possible to align the computational tractability of approximate inference of varying qualities with standard results from algorithmic and computational complexity theory, by combining such an asymptotic analysis with a careful treatment of the variances of all internal Monte Carlo estimators.

This technique opens up other new research opportunities. For example, it may be possible to predict the probable performance of approximate inference by building probabilistic models that use characteristics of problem instances to predict subjective divergences. It may also be possible to use the technique to justify inference heuristics such as \cite{obermeyer2014scaling} and \cite{balakrishnan2006one}, and the stochastic Bayesian relaxations from \cite{DBLP:journals/corr/MansinghkaSJPGT15}, \cite{mansinghka2013approximate}. Finally, it seems fruitful to use the technique to study the query sensitivity of approximate inference \cite{russell_personal_communication}.

Practitioners of probabilistic modeling and inference are all too familiar with the difficulties that come with dependence on approximation algorithms, especially stochastic ones. Diagnosing the convergence of sampling schemes is known to be difficult in theory \cite{diaconis2009markov} and in practice \cite{Cowles1996}. Many practitioners respond by restricting the class of models and queries they will consider. The definition of ``tractable'' is sometimes even taken to be synonymous with ``admits polynomial time algorithms for exactly calculating marginal probabilities'', as in \cite{poon2011sum}. Probabilistic programming throws these difficulties into sharp relief, by making it easy to explore an unbounded space of possible models, queries, and inference strategies. Hardly any probabilistic inference programs come with certificates that they give exact answers in polynomial time.

It is understandable that many practitioners are wary of expressive probabilistic languages. The techniques in this paper make it possible to pursue an alternative approach: use expressive languages for modeling and potentially even also stochastic inference strategies, but also build quantitative models of the time-accuracy profiles of approximate inference, in practice, from empirical data. This is an inherently subjective process, involving qualitative and quantitative assumptions at the meta-level. However, we note that probabilistic programming can potentially help manage this meta-modeling process, providing new probabilistic---or in some sense meta-probabilistic---tools for studying the probable convergence profiles of probabilistic inference programs.

%% file: tex/basic_notation.tex
\section{Basic notation}

The notation $p(z)$ is used to denote the distribution of a random variable, as
well the corresponding probability density function, and we rely on the context to
disambiguate between the two. In particular, the KL divergence from probability
distribution $p(z)$ to probability distribution $q(z)$ is denoted
$\KL(p(z)||q(z))$:
\begin{equation}
\KL(p(z)||q(z)) = \E_{z \sim p(z)} \left[ \log \frac{p(z)}{q(z)} \right]
\end{equation}
where the $p(z)$ and $q(z)$ inside the expectation are density functions which
take values $z$ as input, and $z \sim p(z)$ indicates a random variable with
distribution $p(z)$.  Throughout, when comparing two distributions $p(z)$ and
$q(z)$ we assume that they have equal support ($p(z) = 0 \iff q(z) = 0$).  The
\emph{symmetrized} KL divergence between $p(z)$ and $q(z)$ is
\begin{equation}
\KL(p(z)||q(z)) + \KL(q(z)||p(z))
\end{equation}
In this appendix, we use the shorthand $p$ for $p(z)$, and $\KL(p||q)$ for $\KL(p(z)||q(z))$ when
there is no ambiguity as to the distributions represented by $p$ and $q$.

%% file: tex/deriving.tex
\section{Deriving the subjective divergence}

This section provides a pedagogical derivation of subjective divergence.
Suppose we seek to estimate the KL divergence between two distributions $q(z)$
and $p(z)$ (in this section we do not initially assume these to be approximate
inference or posterior distributions in particular). We walk through a
motivating derivation of the subjective divergence as an approach to this
problem.

\subsection{Monte Carlo estimation}

Suppose we can sample from $q$ and $p$, and that normalized densities of
$q$ and $p$ are available. Then, we can estimate either direction of KL
divergence using simple Monte Carlo, e.g.:
\begin{equation}
\KL(q||p) = \E_{z\sim q} \left[ \log \frac{q(z)}{p(z)} \right] \approx \frac{1}{N} \sum_{i=1}^N \log \frac{q(z_i)}{p(z_i)}
\end{equation}
where $z_i \sim q$. The accuracy of the estimates is determined by the variance
in the log weight ($\log \frac{q(z)}{p(z)}$) and $N$.

\subsection{Symmetrized KL divergence}

Suppose now that only unnormalized densities $\tilde{q}(z)$ and $\tilde{p}(z)$ can
be computed with unknown normalizing constants $Z_P$ and $Z_Q$, but that
we can still sample from $q$ and $p$. Then the two directions of KL divergence
are:
\begin{equation}
\KL(q||p) = \E_{z\sim q} \left[ \log \frac{q(z)}{p(z)} \right] = \E_{z\sim q} \left[ \log \frac{\tilde{q}(z)/Z_Q}{\tilde{p}(z)/Z_P}\right]  = \log \frac{Z_P}{Z_Q} + \E_{z\sim q} \left[ \log \frac{\tilde{q}(z)}{\tilde{p}(z)}\right]  
\end{equation}
\begin{equation}
\KL(p||q) = \E_{z\sim p} \left[ \log \frac{p(z)}{q(z)} \right] = \E_{z\sim p} \left[ \log \frac{\tilde{p}(z)/Z_P}{\tilde{q}(z)/Z_Q} \right] = \log \frac{Z_Q}{Z_P} + \E_{z\sim p} \left[ \log \frac{\tilde{p}(z)}{\tilde{q}(z)}\right]  
\end{equation}
Suppose we can accurately estimate the expectation terms for both of these
quantities using simple Monte Carlo, but that estimating the terms $\log
\frac{Z_P}{Z_Q}$ and $\log \frac{Z_Q}{Z_P}$ is more difficult.

Consider the direction $\KL(q||p)$. Estimating only the expectation term allows
us to estimate \emph{differences} in KL divergence $\KL(q_1||p)$ or
$\KL(q_2||p)$ if the normalizing constants $Z_{Q_1}$ and $Z_{Q_2}$ are the
same. The `evidence lower bound' (ELBO) optimized in variational inference is
such an expectation, in which often $Z_{Q_1} = Z_{Q_2} = 1$. The ELBO is used to
guide a search or optimization process over a space of $q \in \mathcal{Q}$ to
minimize $\KL(q||p)$. However, not knowing the normalizing constant $Z_P$
prevents us from estimating the KL divergence itself.

Note that in the symmetrized KL divergence, the terms containing the
normalizing constants cancel, and we are left with:
\begin{align}
\KL(q||p) + \KL(p||q)
&= \E_{z\sim q} \left[ \log \frac{\tilde{q}(z)}{\tilde{p}(z)} \right] + \E_{z\sim p} \left[ \log \frac{\tilde{p}(z)}{\tilde{q}(z)} \right] \\
&= \E_{z\sim p} \left[ \log \frac{\tilde{p}(z)}{\tilde{q}(z)} \right] - \E_{z\sim q} \left[ \log \frac{\tilde{p}(z)}{\tilde{q}(z)}\right] \\
&= \E_{z\sim p} \left[ \log w(z)\right]  - \E_{z\sim q} \left[ \log w(z)\right]  
\end{align}
where we define the unnormalized weight function as $w(z) :=
\frac{\tilde{p}(z)}{\tilde{q}(z)}$.  Suppose we use a simple Monte Carlo
estimator for each of the two expectations in the above expression of the
symmetric KL divergence by sampling from $q$ and $p$ respectively, and take the
difference in estimates. This can be interpreted as comparing samples from $q$
against samples from $p$ by projecting them through the log-weight function
$\log w(\cdot)$ onto $\mathbb{R}$.

\subsection{Non-oracle reference inference program}

We now refine the setting to more closely match the approximate inference
setting, in which it is relatively easy to sample from $q$, and difficult to
sample from $p$. Specifically, we assume that the term $\E_{z\sim q} \left[ \log w(z)\right]$
is relatively easier to estimate than $\E_{z\sim p} \left[ \log w(z)\right] $. This is often the
case, for example, if $p$ is a posterior distribution and $q$ is the
approximating distribution of a typical inference program. We consider using
samples from a proxy $r(z)$ instead of samples from $p(z)$, for which
$r$ is more efficient to sample from than $p$ itself, but otherwise using
the original weight function $w(z)$ which is defined in terms of $p$ and
$q$. Instead of the symmetric KL divergence between $q$ and $p$ we are then
estimating:
\begin{align}
\E_{z\sim r} \left[\log w(z)\right]  - \E_{z\sim q} \left[ \log w(z)\right]
&= \E_{z\sim r} \left[ \log \frac{\tilde{p}(z)}{\tilde{q}(z)} \right] - \E_{z\sim q} \left[ \log \frac{\tilde{p}(z)}{\tilde{q}(z)}\right]\\
&= \E_{z\sim r} \left[ \log \frac{p(z)}{q(z)} \right] - \E_{z\sim q} \left[ \log \frac{p(z)}{q(z)}\right]\\
&= \E_{z\sim r} \left[ \log \frac{p(z)}{r(z)}\frac{r(z)}{q(z)} \right] - \E_{z\sim q} \left[ \log \frac{p(z)}{q(z)} \right]\\
&= \left( \KL(r||q) - \KL(r||p)\right) + \KL(q||p)
\end{align}
The difference between our expectation and the true symmetrized KL is:
\begin{align}
\KL(r||q) - \KL(r||p) - \KL(p||q)
\end{align}
For $\KL(r||p) = 0$ the difference is zero. Assuming certain conditions on $r$,
we still estimate an upper bound on the symmetrized KL divergence (see
Proposition~\ref{prop-sup:non_oracle_prop_symmetric}) and the KL divergence from
$q(z;x^*)$ to $p(z|x^*)$ (see
Proposition~\ref{prop-sup:non_oracle_prop_forward}).

\subsection{Inference program output marginal density estimators}

We now handle the setting in which the density $q(z)$ is not available, even up
to a normalizing constant, due to the presence of internal random choices $y$
involved in sampling from $q(z)$:
\[
q(z) = \int q(y,z) dy
\]
where $y$ is high-dimensional.
We take $q(y,z)$ to be an inference program, and we refer to $y$ as an inference execution
history. Note that unlike in the main text, the dependence on the data set $x^*$ is omitted in the notation of this section.
When $y$ and $z$ are jointly sampled from $q(y,z)$ by first sampling $y \sim q(y)$ followed by $z|y \sim q(z|y)$, $y$ is the history
of the inference program execution that generated $z$. Consider 
the symmetrized KL divergence:
\begin{align}
\KL(q(z)||p(z)) + \KL(p(z)||q(z))
&= \E_{z\sim p(z)} \left[ \log w(z)\right]  - \E_{z\sim q(z)} \left[ \log w(z) \right]  \\
&= \E_{z\sim p(z)} \left[ \log \frac{\tilde{p}(z)}{q(z)} \right] - \E_{z\sim q(z)} \left[ \log \frac{\tilde{p}(z)}{q(z)} \right]
\end{align}

We will construct a Monte Carlo estimate of the symmetrized KL divergence that
uses estimators $\hat{q}(z)$, which are potentially stochastic given $z$,
instead of the true densities $q(z)$:
\begin{align}
\frac{1}{N} \sum_{i=1}^N \log \frac{\tilde{p}(z^p_i)}{\hat{q}(z^p_i)} - \frac{1}{M} \sum_{j=1}^M \log \frac{\tilde{p}(z^q_j)}{\hat{q}(z^q_j)}
\end{align}
for $z^p_i \sim p(z)$ for $i=1,\ldots,N$ and $z^q_j \sim q(z)$ for $j=1,\ldots,M$.
The expectation of the estimate is:
\begin{align}
\E_{z\sim p(z)} \left[ \E \left[ \log \frac{\tilde{p}(z)}{\hat{q}(z)} \right] \right] - \E_{z \sim q(z)} \left[ \E \left[ \log \frac{\tilde{p}(z)}{\hat{q}(z)} \right] \right]
\end{align}
where the inner expectations are with respect to the distributions of the
random variables $\hat{q}(z)$ conditioned on $z$. We want the expectation of
our estimate to be an upper bound on the true symmetrized KL divergence. To
enforce this, we choose distinct estimators for $\hat{q}(z)$,
denoted $\QIS(z)$ and $\QHM(z)$ respectively, for use with the samples $z \sim p(z)$ and for use with the
samples $z \sim q(z)$ such that the following two conditions hold:
\begin{align}
\E_{z\sim p(z)} \left[ \E \left[ \log \frac{\tilde{p}(z)}{\QIS(z)} \right] \right] \ge \E_{z\sim p(z)} \left[ \log \frac{\tilde{p}(z)}{q(z)} \right]
\end{align}
\begin{align}
\E_{z\sim q(z)} \left[ \E \left[ \log \frac{\tilde{p}(z)}{\QHM(z)} \right] \right] \le \E_{z\sim q(z)} \left[ \log \frac{\tilde{p}(z)}{q(z)} \right]
\end{align}
This will ensure that the expectation of our estimate is greater than the
symmetrized KL divergence. To achieve this, we require that:
\begin{align}
\E \left[ \log \frac{\tilde{p}(z)}{\QIS(z)} \right] \ge  \log \frac{\tilde{p}(z)}{q(z)} \;\;\; \forall z
\end{align}
\begin{align}
\E \left[ \log \frac{\tilde{p}(z)}{\QHM(z)} \right] \le \log \frac{\tilde{p}(z)}{q(z)} \;\;\; \forall z
\end{align}
This is equivalent to the requirement that:
\begin{align}
\E \left[ \log \QIS(z) \right] \le \log q(z) \;\;\; \forall z
\end{align}
\begin{align}
\E \left[ \log \QHM(z) \right] \ge \log q(z) \;\;\; \forall z
\end{align}
As pointed out in \cite{grosse2015sandwiching} (see Lemma~\ref{lemma-sup:estimator_log_bound_unbiased} and Lemma~\ref{lemma-sup:estimator_log_bound_reciprocal_unbiased} in Appendix~\ref{sec:proofs}), these requirements are met if:
\begin{align}
\E \left[ \QIS(z) \right] = q(z) \;\;\; \forall z
\label{eq:unbiased_condition}
\end{align}
\begin{align}
\E \left[ \left( \QHM(z) \right)^{-1} \right] = \left( q(z) \right)^{-1}\;\;\; \forall z
\label{eq:unbiased_reciprocal_condition}
\end{align}
There are potentially many choices for $\QIS(z)$ and $\QHM(z)$ that satisfy these
conditions. To construct the baseline estimators we assume that we can
efficiently compute the joint density $q(y,z)$. For the estimator $\QIS(z)$ we use
an importance sampling estimator with importance distribution $m(y;z)$ where
\begin{align}
\E_{y \sim m(y;z)} \left[ \frac{q(y,z)}{m(y;z)} \right] = q(z) \;\;\; \forall z
\end{align}
Defining:
\begin{equation}
\QIS(z) = \frac{1}{L} \sum_{\ell=1}^L\frac{q(y_{\ell},z)}{m(y_{\ell};z)} \;\;\; \mbox{ for } \;\;\; y_{\ell} \sim m(y;z)
\end{equation}
satisfies the unbiasedness condition of Equation~\ref{eq:unbiased_condition} for $\QIS(z)$. 
To construct the estimator $\QHM(z)$ we
note that
\begin{align}
\E_{y \sim q(y|z)} \left[ \frac{m(y;z)}{q(y,z)} \right] = \frac{1}{q(z)} \;\;\; \forall z
\end{align}
We define $\QHM(z)$ as a harmonic mean estimator:
\begin{equation}
\QHM(z) = \frac{L}{\sum_{\ell=1}^L \frac{m(y_{\ell};z)}{q(y_{\ell},z)}} \;\;\; \mbox{ for } \;\;\; y_{\ell} \sim q(y|z)
\end{equation}
which satisfies the unbiased reciprocal condition of Equation~\ref{eq:unbiased_reciprocal_condition} for $\QHM(z)$. Algorithm~\ref{alg:general} uses $L = 1$ for both $\QIS(z)$ and $\QHM(z)$, and obtains the sample of inference program
execution history $y \sim q(y|z)$ from the joint sample that generated $z$.
Note that only one such sample is immediately available for each $z$, although
we could conceivably start a Markov chain at the exact sample $y$ with $q(y|z)$
as its stationary distribution to obtain more samples $y$ marginally
distributed according to $q(y|z)$. Using more sophisticated versions of $\QHM(z)$
and $\QIS(z)$ is left for future work.  Note that for the single-particle
baseline estimators and an oracle reference, the sole determiner of the gap between the subjective divergence and the symmetrized KL is the quality of
the distribution $m(y;z)$ as an approximation to $q(y|z)$. We refer to $m(y;z)$
as the meta-inference distribution.

%% file: tex/proofs.tex
\section{Proofs}
\label{sec:proofs}

\begin{frm-lemma-sup}
For $\QHM(z;x^*)$ such that $\E\left[\left(\QHM(z;x^*)\right)^{-1}\right] = q(z;x^*)^{-1}$ for all $z$,\\
$\E_{z\sim q(z;x^*)} \left[ \E \left[ \log \frac{p(z,x^*)}{\QHM(z;x^*)} \right] \right] \le \log p(x^*) - \KL(q(z;x^*)||p(z|x^*))$
\label{lemma-sup:inference_program_bound}
\end{frm-lemma-sup}
\begin{proof}
\begin{align}
\shortintertext{Factoring out the normalizing constant $p(x^*)$ using $p(z,x^*)=p(z|x^*)p(x^*)$:}
\E_{z\sim q(z;x^*)} \left[ \E \left[ \log \frac{p(z,x^*)}{\QHM(z;x^*)} \right] \right]
&= \E_{z\sim q(z;x^*)} \left[ \E \left[ \log \frac{p(z|x^*) p(x^*)}{\QHM(z;x^*)} \right] \right]\\
&= \E_{z\sim q(z;x^*)} \left[ \E \left[ \log p(x^*) + \log \frac{p(z|x^*)}{\QHM(z;x^*)} \right] \right]\\
\shortintertext{Linearity of expectation:}
&= \E_{z\sim q(z;x^*)} \left[ \E \left[ \log p(x^*) \right] \right] + \E_{z\sim q(z;x^*)} \left[ \E \left[ \log \frac{p(z|x^*)}{\QHM(z;x^*)} \right] \right]\\
\shortintertext{The normalizing constant $p(x^*)$ is a constant:}
&= \log p(x^*) + \E_{z\sim q(z;x^*)} \left[ \E \left[ \log \frac{p(z|x^*)}{\QHM(z;x^*)} \right] \right]\\
&= \log p(x^*) + \E_{z\sim q(z;x^*)} \left[ \E \left[ \log p(z|x^*) - \log \QHM(z;x^*) \right] \right]\\
\shortintertext{Linearity of expectation:}
&= \log p(x^*) + \E_{z\sim q(z;x^*)} \left[ \E \left[ \log p(z|x^*) \right] - \E \left[ \log \QHM(z;x^*) \right] \right]\\
\shortintertext{Conditioned on $z$, $p(z|x^*)$ is a constant:}
&= \log p(x^*) + \E_{z\sim q(z;x^*)} \left[ \log p(z|x^*) - \E \left[ \log \QHM(z;x^*) \right] \right]\\
\shortintertext{Using Lemma~\ref{lemma-sup:estimator_log_bound_reciprocal_unbiased} (see below) with the given condition $\E\left[\left(\QHM(z;x^*)\right)^{-1}\right] = q(z;x^*)^{-1}$ for all $z$:}
&\le \log p(x^*) + \E_{z\sim q(z;x^*)} \left[ \log p(z|x^*) - \log q(z;x^*)  \right]\\
&= \log p(x^*) + \E_{z\sim q(z;x^*)} \left[ \log \frac{p(z|x^*)}{q(z;x^*)} \right]\\
&= \log p(x^*) - \E_{z\sim q(z;x^*)} \left[ \log \frac{q(z;x^*)}{p(z|x^*)} \right]\\
\shortintertext{Using the definition of Kullback-Leibler (KL) divergence \cite{cover2012elements}:}
&= \log p(x^*) - \KL(q(z;x^*)||p(z|x^*))
\end{align}
\end{proof}

\begin{frm-lemma-sup}
For $\QIS(z;x^*)$ such that $\E[\QIS(z;x^*)] = q(z;x^*)$ for all $z$,\\
$\E_{z\sim p(z|x^*)} \left[ \E \left[ \log \frac{p(z,x^*)}{\QIS(z;x^*)} \right] \right] \ge \log p(x^*) + \KL(p(z|x^*)||q(z;x^*))$
\label{lemma-sup:reference_program_bound}
\end{frm-lemma-sup}
\begin{proof}
\begin{align}
\shortintertext{Factoring out the normalizing constant $p(x^*)$ using $p(z,x^*)=p(z|x^*)p(x^*)$:}
\E_{z\sim p(z|x^*)} \left[ \E \left[ \log \frac{p(z,x^*)}{\QIS(z;x^*)} \right] \right]
&= \E_{z\sim p(z|x^*)} \left[ \E \left[ \log \frac{p(z|x^*) p(x^*)}{\QIS(z;x^*)} \right] \right]\\
&= \E_{z\sim p(z|x^*)} \left[ \E \left[ \log p(x^*) + \log \frac{p(z|x^*)}{\QIS(z;x^*)} \right] \right]\\
\shortintertext{Linearity of expectation:}
&= \E_{z\sim p(z|x^*)} \left[ \E \left[ \log p(x^*) \right] \right] + \E_{z\sim p(z|x^*)} \left[ \E \left[ \log \frac{p(z|x^*)}{\QIS(z;x^*)} \right] \right]\\
\shortintertext{The normalizing constant $p(x^*)$ is a constant:}
&= \log p(x^*) + \E_{z\sim p(z|x^*)} \left[ \E \left[ \log \frac{p(z|x^*)}{\QIS(z;x^*)} \right] \right]\\
&= \log p(x^*) + \E_{z\sim p(z|x^*)} \left[ \E \left[ \log p(z|x^*) - \log \QIS(z;x^*) \right] \right]\\
\shortintertext{Linearity of expectation:}
&= \log p(x^*) + \E_{z\sim p(z|x^*)} \left[ \E \left[ \log p(z|x^*) \right] - \E \left[ \log \QIS(z;x^*) \right] \right]\\
\shortintertext{Conditioned on $z$, $p(z|x^*)$ is a constant:}
&= \log p(x^*) + \E_{z\sim p(z|x^*)} \left[ \log p(z|x^*) - \E \left[ \log \QIS(z;x^*) \right] \right]\\
\shortintertext{Using Lemma~\ref{lemma-sup:estimator_log_bound_unbiased} (see below) with the given condition $\E[\QIS(z;x^*)] = q(z;x^*)$ for all $z$:}
&\ge \log p(x^*) + \E_{z\sim p(z|x^*)} \left[ \log p(z|x^*) - \log q(z;x^*) \right]\\
&= \log p(x^*) + \E_{z\sim p(z|x^*)} \left[ \log \frac{p(z|x^*)}{q(z;x^*)} \right]\\
\shortintertext{Using the definition of Kullback-Leibler (KL) divergence \cite{cover2012elements}:}
&= \log p(x^*) + \KL(p(z|x^*)||q(z;x^*))
\end{align}
\end{proof}

\begin{frm-lemma-sup}[Unbiased estimators are lower bound log estimators \cite{grosse2015sandwiching}]
For any $\hat{x}$ such that $\E[\hat{x}] = x$, $\E[\log \hat{x}] \le \log x$
\label{lemma-sup:estimator_log_bound_unbiased}
\end{frm-lemma-sup}
\begin{proof}
\begin{align}
\shortintertext{By Jensen's inequality, since $\log(\cdot)$ is concave:}
\E[\log \hat{x}] &\le \log \E[\hat{x}]\\
\shortintertext{By given condition $\E[\hat{x}] = x$:}
&= \log x
\end{align}
\end{proof}

\begin{frm-lemma-sup}[Unbiased reciprocal estimators are upper bound log estimators \cite{grosse2015sandwiching}]
For any $\hat{x}$ such that $\E[\left(\hat{x}\right)^{-1}] = x^{-1}$, $\E[\log \hat{x}] \ge \log x$
\label{lemma-sup:estimator_log_bound_reciprocal_unbiased}
\end{frm-lemma-sup}
\begin{proof}
\begin{align}
\E[\log \hat{x}] &= \E\left[ -\log \left( \frac{1}{\hat{x}} \right)\right]\\
\shortintertext{By Jensen's inequality, since $-\log(\cdot)$ is convex:}
&\ge -\log \left( \E \left[ \frac{1}{\hat{x}}\right]\right)\\
\shortintertext{By given condition $\E[\left(\hat{x}\right)^{-1}] = x^{-1}$:}
&= - \log \left( \frac{1}{x}\right)\\
&= \log x
\end{align}
\end{proof}

\begin{frm-lemma-sup}
For $\QIS(z)$ such that $\E[\QIS(z;x^*)] = q(z;x^*)$,\\
$\E_{z\sim r(z;x^*)} \left[ \E \left[ \log \frac{p(z,x^*)}{\QIS(z;x^*)} \right] \right] \ge \log p(x^*) - \KL(r(z;x^*)||p(z|x^*)) + \KL(r(z;x^*)||q(z;x^*))$
\label{lemma-sup:non_oracle_reference}
\end{frm-lemma-sup}
\begin{proof}~
\begin{align}
&\E_{z\sim r(z;x^*)} \left[ \E \left[ \log \frac{p(z,x^*)}{\QIS(z;x^*)} \right] \right]\\
&\;\;\;= \log p(x^*) + \E_{z\sim r(z;x^*)} \left[ \E \left[ \log \frac{p(z|x^*)}{\QIS(z;x^*)} \right] \right]\\
&\;\;\;= \log p(x^*) + \E_{z\sim r(z;x^*)} \left[ \log p(z|x^*) - \E \left[ \log \QIS(z;x^*) \right] \right]\\
&\;\;\;\ge \log p(x^*) + \E_{z\sim r(z;x^*)} \left[ \log p(z|x^*) - \log q(z;x^*) \right]\\
&\;\;\;= \log p(x^*) + \E_{z\sim r(z;x^*)} \left[ \log \frac{p(z|x^*)}{q(z;x^*)} \right]\\
&\;\;\;= \log p(x^*) + \E_{z\sim r(z;x^*)} \left[ \log \frac{p(z|x^*)}{r(z;x^*)} \frac{r(z;x^*)}{q(z;x^*)} \right]\\
&\;\;\;= \log p(x^*) + \E_{z\sim r(z;x^*)} \left[ \log \frac{p(z|x^*)}{r(z;x^*)} \right] + \E_{z\sim r(z;x^*)} \left[ \log \frac{r(z;x^*)}{q(z;x^*)} \right]\\
&\;\;\;= \log p(x^*) - \KL(r(z;x^*)||p(z|x^*)) + \KL(r(z;x^*)||q(z;x^*))
\end{align}
\end{proof}

\begin{frm-prop-sup} If an oracle reference program is used then\\
$\SBJ(q(z;x^*)||p(z|x^*)) \ge \KL(q(z;x^*)||p(z|x^*)) + \KL(p(z|x^*)||q(z;x^*))$
\label{frm-prop-sup:symmetric_bound}
\end{frm-prop-sup}
\begin{proof}
\begin{align}
\shortintertext{The definition of subjective divergence:}
&\SBJ(q(z;x^*)||p(z|x^*))\\
&\;\;\;:= \E_{z \sim r(z;x^*)} \left[ \E \left[\log \frac{p(z,x^*)}{\QIS(z;x^*)} \right] \right] - \E_{z\sim q(z;x^*)} \left[ \E \left[\log \frac{p(z,x^*)}{\QHM(z;x^*)}\right] \right]\\
\shortintertext{Using an oracle reference inference program ($r(z;x^*) = p(z|x^*) \;\;\forall z$):}
&\;\;\;= \E_{z \sim p(z|x^*)} \left[ \E \left[\log \frac{p(z,x^*)}{\QIS(z;x^*)} \right] \right] - \E_{z\sim q(z;x^*)} \left[ \E \left[\log \frac{p(z,x^*)}{\QHM(z;x^*)}\right] \right]\\
\shortintertext{Using Lemma~\ref{lemma-sup:inference_program_bound} to bound the second expectation:}
&\;\;\;\ge \E_{z \sim p(z|x^*)} \left[ \E \left[\log \frac{p(z,x^*)}{\QIS(z;x^*)} \right] \right] - \left( \log p(x^*) - \KL(q(z;x^*)||p(z|x^*)) \right)\\
\shortintertext{Using Lemma~\ref{lemma-sup:reference_program_bound} to bound the first expectation:}
&\;\;\;\ge \left( \log p(x^*) + \KL(p(z|x^*)||q(z;x^*))\right) - \left( \log p(x^*) - \KL(q(z;x^*)||p(z|x^*)) \right)\\
\shortintertext{The log normalizing constant $\log p(x^*)$ cancels:}
&\;\;\;= \KL(p(z|x^*)||q(z;x^*)) + \KL(q(z;x^*)||p(z|x^*))
\end{align}
\end{proof}

\begin{frm-prop-sup} If $\KL(r(z;x^*)||p(z|x^*)) \le \KL(r(z;x^*)||q(z;x^*)) - \KL(p(z|x^*)||q(z;x^*))$
then $\SBJ(q(z;x^*)||p(z|x^*)) \ge \KL(q(z;x^*)||p(z|x^*)) + \KL(p(z|x^*)||q(z;x^*))$
\label{prop-sup:non_oracle_prop_symmetric}
\end{frm-prop-sup}
\begin{proof}
Taking the definition of subjective divergence and the difference of the bounds of Lemma~\ref{lemma-sup:non_oracle_reference} and Lemma~\ref{lemma-sup:inference_program_bound} gives
\begin{align}
\SBJ(q(z;x^*)||p(z|x^*)) &\ge -\KL(r(z;x^*)||p(z|x^*)) + \KL(r(z;x^*)||q(z;x^*)) + \KL(q(z;x^*)||p(z|x^*))
\end{align}
If $\KL(r(z;x^*)||p(z|x^*)) \le \KL(r(z;x^*)||q(z;x^*)) - \KL(p(z|x^*)||q(z;x^*))$ then
\[
\KL(p(z|x^*)||q(z;x^*)) \le \KL(r(z;x^*)||q(z;x^*)) - \KL(r(z;x^*)||p(z|x^*))
\]
\[
\SBJ(q(z;x^*)||p(z|x^*)) \ge \KL(p(z|x^*)||q(z;x^*)) + \KL(q(z;x^*)||p(z|x^*))
\]
\end{proof}

\begin{frm-prop-sup} If $\KL(r(z;x^*)||p(z|x^*)) \le \KL(r(z;x^*)||q(z;x^*))$ then
\begin{align*}
\E_{z\sim r(z;x^*)} \left[ \E \left[ \log \frac{p(z,x^*)}{\QIS(z;x^*)} \right] \right] &\ge \log p(x^*)\\
\SBJ(q(z;x^*)||p(z|x^*)) &\ge \KL(q(z;x^*)||p(z|x^*))
\end{align*}
\label{prop-sup:non_oracle_prop_forward}
\end{frm-prop-sup}
\begin{proof}
The first result follows from Lemma~\ref{lemma-sup:non_oracle_reference}.
The second result follows from the definition of subjective divergence and
the difference in the bounds of the first result and
Lemma~\ref{lemma-sup:inference_program_bound}.
\end{proof}

%% file: tex/metainference_analysis.tex
\section{Effect of quality of meta-inference program}
\label{sec:metainference_analysis}

This section analyzes the difference between subjective divergence and the
symmetrized KL divergence for the procedure of Algorithm~\ref{alg:general} in
the oracle reference setting. In this case, the gap between the subjective
divergence and the true KL divergence is the symmetrized conditional relative
entropy between the meta-inference distribution and the conditional distribution on
execution histories given inference program output. To see this, first consider
the expected log estimated weight under the inference program:
\begin{align}
&\E_{z\sim q(z;x^*)} \left[ \E \left[ \log \frac{p(z,x^*)}{\QHM(z;x^*)} \right] \right]\\
&= \E_{z\sim q(z;x^*)}\left[ \E_{y|z \sim q(y|z;x^*)} \left[ \log \frac{p(z|x^*)m(y;z,x^*)}{q(y,z;x^*)} \right] \right] + \log p(x^*)\\
&= \log p(x^*) - \KL(q(y,z;x^*)||p(z|x^*)m(y;z,x^*))\\
\shortintertext{Using the chain rule for joint KL divergence \cite{cover2012elements}:}
&= \log p(x^*) - \KL(q(z;x^*)||p(z|x^*)) - \E_{z \sim q(z;x^*)} \left[ \KL(q(y|z;x^*)||m(y;z,x^*)) \right]
\end{align}
Next, consider the expected log estimated weight under the under the oracle
reference program:
\begin{align}
&\E_{z\sim p(z|x^*)} \left[ \E \left[ \log \frac{p(z,x^*)}{\QIS(z;x^*)} \right] \right]\\
&= \E_{z\sim p(z|x^*)} \left[ \E_{y|z \sim m(y;z,x^*)} \left[ \log \frac{p(z|x^*)m(y;z,x^*)}{q(y,z;x^*)} \right] \right] + \log p(x^*)\\
&= \log p(x^*) + \KL(p(z|x^*)m(y;z,x^*)||q(y,z;x^*))\\
\shortintertext{Using the chain rule for joint KL divergence \cite{cover2012elements}:}
&= \log p(x^*) + \KL(p(z|x^*)||q(z;x^*)) + \E_{z \sim p(z|x^*)} \left[ \KL(m(y;z,x^*)||q(y|z;x^*)) \right]
\end{align}
The difference in these expectations is the subjective divergence $\SBJ$:
\begin{align}
&\SBJ(q(z;x^*)||p(z|x^*))\\
&= \E_{z\sim p(z|x^*)} \left[ \E \left[ \log \frac{p(z,x^*)}{\QIS(z;x^*)} \right] \right] - \E_{z\sim q(z;x^*)} \left[ \E \left[ \log \frac{p(z,x^*)}{\QHM(z;x^*)} \right] \right]\\
&= \left( \KL(p(z|x^*)||q(z;x^*)) + \KL(q(z;x^*)||p(z|x^*))\right)\\
&\;\;\; + \E_{z \sim p(z|x^*)} \left[ \KL(m(y;z,x^*)||q(y|z;x^*)) \right] + \E_{z \sim q(z;x^*)} \left[ \KL(q(y|z;x^*)||m(y;z,x^*)) \right]
\end{align}
Therefore the looseness of the bound on the actual symmetric KL divergence is:
\begin{equation}
\E_{z \sim p(z|x^*)} \left[ \KL(m(y;z,x^*)||q(y|z;x^*)) \right] + \E_{z \sim q(z;x^*)} \left[ \KL(q(y|z;x^*)||m(y;z,x^*)) \right]
\end{equation}
To gain intuition about how the gap is related to the accuracy of inference
output marginal density estimation, consider the variance of $\QIS(z;x^*)$ and the bias of the induced estimator of $\log q(z;x^*)$:
\begin{align}
\Var\left( \frac{\QIS(z;x^*)}{q(z;x^*)} \right)
&= \E \left[ \left( \frac{\QIS(z;x^*)}{q(z;x^*)} \right)^2 - \left(\E\left[ \frac{\QIS(z;x^*)}{q(z;x^*)} \right]\right)^2 \right]\\
&= \E_{y|z \sim m(y;z,x^*)} \left[ \left( \frac{q(y,z;x^*)}{q(z;x^*) m(y;z,x^*)} \right)^2 - 1 \right]\\
&= \E_{y|z \sim m(y;z,x^*)} \left[ \left( \frac{q(y|z;x^*)}{m(y;z,x^*)} \right)^2 - 1 \right]\\
&= \chi_P^2 (m(y;z,x^*)||q(y|z;x^*))\\
\log q(z;x^*) - \E \left[ \log \QIS(z;x^*) \right]
&= \E \left[ \log \frac{q(z;x^*)}{\QIS(z;x^*)} \right]\\
&= \E_{y|z \sim m(y;z,x^*)} \left[ \log \frac{q(z;x^*) m(y;z,x^*)}{q(y,z;x^*)} \right]\\
&= \E_{y|z \sim m(y;z,x^*)} \left[ \log \frac{m(y;z,x^*)}{q(y|z;x^*)} \right]\\
&= \KL(m(y;z,x^*)||q(y|z;x^*))
\end{align}
Also consider the variance of $\left(\QHM(z;x^*)\right)^{-1}$ and the bias of the induced
estimator for $\log q(z;x^*)$:
\begin{align}
\Var\left( \frac{q(z;x^*)}{\QHM(z;x^*)} \right)
&= \E \left[ \left( \frac{q(z;x^*)}{\QHM(z;x^*)} \right)^2 - \left(\E\left[ \frac{q(z;x^*)}{\QHM(z;x^*)} \right]\right)^2 \right]\\
&= \E_{y|z \sim q(y|z;x^*)} \left[ \left( \frac{q(z;x^*)m(y;z,x^*)}{q(y,z;x^*)} \right)^2 - 1 \right]\\
&= \E_{y|z \sim q(y|z;x^*)} \left[ \left( \frac{m(y;z,x^*)}{q(y|z;x^*)} \right)^2 - 1 \right]\\
&= \chi_P^2 (q(y|z;x^*)||m(y;z,x^*))\\
\E \left[ \log \QHM(z;x^*) \right] - \log q(z;x^*)
&= \E \left[ \log \frac{\QHM(z;x^*)}{q(z;x^*)} \right]\\
&= \E_{y|z \sim q(y|z;x^*)} \left[ \log \frac{q(y,z;x^*)}{q(z;x^*) m(y;z,x^*)} \right]\\
&= \E_{y|z \sim q(y|z;x^*)} \left[ \log \frac{q(y|z;x^*)}{m(y;z,x^*)} \right]\\
&= \KL(q(y|z;x^*)||m(y;z,x^*))
\end{align}
Above, $\chi_P^2$ is the Pearson chi-square divergence
\cite{nielsen2013chi}:
\begin{align}
\chi_P^2(p(y)||q(y))
&:= \int \frac{\left( q(y) - p(y)\right)^2}{p(y)} dy\\
&= \int p(y) \frac{\left( q(y) - p(y)\right)^2}{p(y)^2} dy\\
&= \int p(y) \frac{q(y)^2 + p(y)^2 - 2 p(y) q(y)}{p(y)^2} dy\\
&= \int p(y) \left( \left(\frac{q(y)}{p(y)}\right)^2 - 1\right) dy
\end{align}

%% file: tex/application_derivations.tex
\section{Derivations for specific inference programs}
\label{sec:app_derivations}

We now show how Algorithm~\ref{alg:general} can be applied to estimate
subjective divergences for three large classes of approximate inference
programs: ``assessable'' inference, sequential stochastic approximate
inference, and particle filtering in state space models.

For convenience, we first introduce new notation specific to the baseline
inference output marginal density estimators $\QIS$ and $\QHM$ that are used in
Algorithm~\ref{alg:general}. Since in this setting, both $\QIS$ and $\QHM$
involve sampling a single inference execution history $y$, and returning an
estimate $q(y,z;x^*)/m(y;z,x^*)$, we denote the estimated weight for a latent
sample $z$, conditioned on a sampled inference execution history $y$, as:
\begin{equation}
\hat{w}_y(z) := \frac{p(z,x^*)}{\left( \frac{q(y,z;x^*)}{m(y;z,x^*)} \right)} = \frac{p(z,x^*) m(y;z,x^*)}{q(y,z;x^*)} 
\end{equation}
In order to use Algorithm~\ref{alg:general}, we must be able to efficiently
compute the function $\hat{w}_y(z)$ and sample from the meta-inference program
$m(y;z,x^*)$. This section lists constructions of $q(y,z;x^*)$ and $m(y;z,x^*)$
that satisfy these properties.

\subsection{Assessable inference}
If the density $q(z;x^*)$ can be efficiently computed exactly, we consider
$q(z;x^*)$ an \emph{assessable inference program}. Inference output marginal
density estimators and meta-inference are not required to estimate subjective
divergence for assessable inference programs, and the procedure of
Algorithm~\ref{alg:general} can be simplified to Algorithm~\ref{alg:assessable}.
Examples of assessable inference include simple variational families for which the density of the
variational approximation, $q_{\theta}(z;x^*)$ where $\theta$ are the variational
parameters, can be efficiently computed.
\begin{algorithm}
    \caption{Subjective divergence estimation for assessable inference programs}
    \label{alg:assessable}
    \begin{algorithmic}[1]
        \Require{Assessable inference program $z|x \sim q(z;x^*)$, reference inference program $z|x \sim r(z;x^*)$, number of reference replicates $N$, number of inference replicates $M$}
        \For{$i \gets 1 \textrm{ to } N$}
            \Sample{$z^r_i$}{$r(z;x^*)$}
            \Let{$w^r_i$}{$\frac{p(z^r_i,x^*)}{q(z^r_i;x^*)}$}
        \EndFor
        \For{$j \gets 1 \textrm{ to } M$}
            \Sample{$z^q_j$}{$q(z;x^*)$}
            \Let{$w^q_j$}{$\frac{p(z^q_j,x^*)}{q(z^q_j;x^*)}$}
        \EndFor
        \State \Return{$\frac{1}{N}\sum_{i=1}^N \log w^r_i - \frac{1}{M} \sum_{j=1}^M \log w^q_j$}
    \end{algorithmic}
\end{algorithm}

\subsection{Sequential stochastic approximate inference programs}
\label{sec:sequential_inference}
Consider a sequential stochastic inference program that proceeds through a
series of steps with intermediate internal states $y_1 \in \mathcal{Y}_1,
\ldots, y_{T} \in \mathcal{Y}_{T}$ and returns a final state $z \in
\mathcal{Z}$, such that the joint distribution of the inference program at this
level of representation factorizes into a Markov chain:
\begin{equation} \label{eq:sai_representation}
q(y,z;x^*) = q(y_1;x^*) \left[ \prod_{t=1}^{T-1} q(y_{t+1}|y_t;x^*) \right] q(z|y_{T};x^*)
\end{equation}
In general the intermediate steps $y_t$ need not share common state spaces
$\mathcal{Y}_t$.  The approximating distribution of the inference program is
defined as the marginal distribution of its output: $q(z;x^*)$. Note that evaluating the
density $q(z;x^*)$ is generally computationally intractable. The optimal
meta-inference distribution for this representation also factorizes into a
Markov chain:
\begin{equation} \label{eq:sai_optimal_metainference}
m^*(y;z,x^*) = q(y_{1:T}|z;x^*) = \left[ \prod_{t=1}^{T-1} q(y_{t}|y_{t+1};x^*) \right] q(y_T|z;x^*)
\end{equation}
Although it may be difficult to construct efficient programs which sample from
the optimal meta-inference distribution,
Equation~\ref{eq:sai_optimal_metainference} suggests that we can start by
designing meta-inference programs that sample states $y_t$ in reverse according
to a Markov chain:
\begin{equation}
m(y;z,x^*) = \left[\prod_{t=1}^{T-1} m(y_t|y_{t+1};z,x^*) \right] m(y_T;z,x^*)
\end{equation}
This mirrors the construction used in \cite{DBLP:conf/icml/SalimansKW15} to estimate
variational lower bounds for Markov chain Monte Carlo. The variational lower
bound of \cite{DBLP:conf/icml/SalimansKW15} corresponds to the inference program term
in subjective divergence with the baseline meta-inference estimator $\QHM$:
\begin{align}
\E_{z\sim q(z;x^*)} \left[ \E \left[\log \frac{p(z,x^*)}{\QHM(z;x^*)}\right] \right]
&= \E_{z\sim q(z;x^*)} \left[ \E_{y|z \sim q(y|z;x^*)} \left[ \log \frac{p(z,x^*) m(y;z,x^*)}{q(y,z;x^*)} \right] \right]
\end{align}
We next derive and analyze meta-inference programs for two instances of
sequential stochastic approximate inference.

\subsubsection{Detailed balance transitions with state extensions}

The derivation of this section uses an inference program corresponding to the
single particle version of Algorithm~2 of \cite{grosse2015sandwiching} and a
meta-inference program corresponding to the single particle version of
Algorithm~3 of \cite{grosse2015sandwiching}.

Suppose that the internal states $y_t$ are defined on state spaces of
increasing dimension. In particular, suppose each intermediate state $y_t$ for
$t=2,\ldots,T$ decomposes into two components $y_t = (u_{t-1}, v_t)$, and $y_1
= v_1$, where $v_t \in \mathcal{V}_t$ for $t=1,\ldots,T$ and $u_t \in
\mathcal{U}_t = \mathcal{U}_{t-1} \times \mathcal{V}_t$ for $t=2,\ldots,{T-1}$
and $\mathcal{U}_1 = \mathcal{V}_1$, and $z = u_T \in \mathcal{Z} = \mathcal{U}_T = \mathcal{U}_{T-1}
\times \mathcal{V}_T$. The inference program is composed of a sequence of
extension steps $q(v_t|u_{t-1};x^*)$ and transition steps $q(u_t|u_{t-1},v_t;x^*)
= k_t(u_t;u_{t-1},v_t)$, and the joint density is:
\begin{equation} \label{eq:sai_dbse_inference_program}
q(y,z;x^*) = q(v_1;x^*) k_1(u_1;v_1) \left[ \prod_{t=2}^{T-1} q(v_t|u_{t-1};x^*) k_t(u_t;u_{t-1},v_t) \right] q(v_T|u_{T-1};x^*) k_T(z;u_{T-1},v_T)
\end{equation}
We assume that each transition operator $k_t$ satisfies the detailed balance
condition for some target distribution $p_t$ defined on $\mathcal{U}_t$ such
that the final target distribution is the posterior ($p_T(z) = p(z|x^*)$):
\begin{equation}
p_t(u) k_t(u';u) = p_t(u') k_t(u;u') \;\; \forall u, u' \in \mathcal{U}_t, t=1,\ldots,T
\end{equation}
Consider the conditional distributions that comprise the optimal meta-inference
Markov chain of Equation~\ref{eq:sai_optimal_metainference} for this setting:
\begin{align}
m^*(y_t|y_{t+1};z,x^*) = q(y_t|y_{t+1};x^*)
&= q(u_{t-1},v_t|u_t,v_{t+1};x^*)\\
&= q(u_{t-1},v_t|u_t;x^*)\\
&= \frac{q(u_{t-1},v_t,u_t;x^*)}{q(u_t;x^*)}\\
&= \frac{q(u_{t-1},v_t;x^*)}{q(u_t;x^*)} k_t(u_t;u_{t-1},v_t)
\end{align}
To derive a meta-inference program we approximate the optimal conditionals with:
\begin{equation}
m(y_t|y_{t+1};z,x^*) = m(u_{t-1},v_t|u_t;z,x^*) = \frac{p_t(u_{t-1}, v_t)}{p_t(u_t)} k_t(u_t; u_{t-1}, v_t) = k_t(u_{t-1}, v_t; u_t)
\end{equation}
Assuming that $q(u_t;x^*) = p_t(u_t)$ amounts to assuming that the operator $k_t$
converges to $p_t$ and assuming that $q(u_{t-1},v_t;x^*) = p_t(u_{t-1},v_t)$ amounts
to assuming that the operator $k_{t-1}$ converges to $p_{t-1}$ and that
$p_{t-1}(u_{t-1}) q(v_t| u_{t-1};x^*) = p_t(u_{t-1},v_t)$. Composing these conditional
distributions, the full meta-inference program consists of running the
transition operators $k_t$ in reverse order:
\begin{equation} \label{eq:sai_metainference}
m(y;z,x^*) = k_1(v_1;u_1) \left[ \prod_{t=2}^{T-1} k_t(u_{t-1},v_t;u_t) \right] k_T(u_{T-1},v_T;z)
\end{equation}
We define $\tilde{p}_t$ as an unnormalized density for target distribution
$p_t$ with arbitrary normalizing constant for $t=1,\ldots,T-1$, except for
$p_T$, for which the unnormalized density is defined as $\tilde{p}_T(z) :=
p(z, x^*)$ with normalizing constant $p(x^*)$.
The weight estimate for the meta-inference program is then:
\begin{align}
&\hat{w}_y(z) =\frac{p(z,x^*)m(y;z,x^*)}{q(y,z;x^*)}\\
&\;=\frac{p(z,x^*) k_1(v_1;u_1) \left[ \prod_{t=2}^{T-1} k_t(u_{t-1},v_t;u_t)\right] k_T(u_{T-1},v_T;z)}{q(v_1;x^*) k_1(u_1;v_1) \left[ \prod_{t=2}^{T-1} q(v_t|u_{t-1};x^*) k_t(u_t;u_{t-1},v_t)\right] q(v_T|u_{T-1};x^*) k_T(z;u_{T-1},v_T)}\\
&\;=\frac{p(z,x^*)}{q(v_1;x^*)\prod_{t=2}^{T}q(v_t|u_{t-1};x^*)} \frac{k_1(v_1;u_1)}{k_1(u_1;v_1)} \left[ \prod_{t=2}^{T-1} \frac{k_t(u_{t-1},v_t;u_t)}{k_t(u_t;u_{t-1},v_t)} \right] \frac{k_T(u_{T-1},v_T;z)}{k_T(z;u_{T-1},v_T)}\\
&\;=\frac{p_T(z) p(x^*)}{q(v_1;x^*) \prod_{t=2}^T q(v_t|u_{t-1};x^*)} \frac{p_1(v_1)}{p_1(u_1)} \left[ \prod_{t=2}^{T-1} \frac{p_t(u_{t-1},v_t)}{p_t(u_t)} \right] \frac{p_T(u_{T-1},v_T)}{p_T(z)}\\
&\;=\frac{p_T(z)}{q(v_1;x^*) \prod_{t=2}^T q(v_t|u_{t-1};x^*)} \frac{p_1(v_1)}{p_1(u_1)} \left[ \prod_{t=2}^{T-1} \frac{p_t(u_{t-1},v_t)}{p_t(u_t)} \right] \frac{\tilde{p}_T(u_{T-1},v_T)}{p_T(z)}\\
&\;=\frac{1}{q(v_1;x^*) \prod_{t=2}^T q(v_t|u_{t-1};x^*)} \frac{p_1(v_1)}{p_1(u_1)} \left[ \prod_{t=2}^{T-1} \frac{p_t(u_{t-1},v_t)}{p_t(u_t)} \right] \frac{\tilde{p}_T(u_{T-1},v_T)}{1}\\
&\;=\frac{1}{q(v_1;x^*) \prod_{t=2}^T q(v_t|u_{t-1};x^*)}\frac{\tilde{p}_1(v_1)}{\tilde{p}_1(u_1)} \frac{\tilde{p}_2(u_1,v_2)}{\tilde{p}_2(u_2)} \cdots \frac{\tilde{p}_{T-1}(u_{T-2},v_{T-1})}{\tilde{p}_{T-1}(u_{T-1})} \frac{\tilde{p}_T(u_{T-1},v_T)}{1} \\
&\;=\frac{\tilde{p}_1(v_1)}{q(v_1;x^*)} \prod_{t=1}^{T-1} \frac{\tilde{p}_{t+1}(u_t,v_{t+1})}{\tilde{p}_t(u_t) q(v_{t+1}|u_t;x^*)}
\end{align}

\subsubsection{Coarse representation of inference programs}

Significantly, each of the operators $k_t$ may be composition of a large
number of steps of primitive transition operators satisfying detailed balance
(e.g. Metropolis Hastings kernels) for target distribution $p_t$. Also, each MH
operator may contain additional random choices such as accept and reject decisions.
The execution histories $y$ of $q(y,z;x^*)$ in Equation~\ref{eq:sai_dbse_inference_program} do not
represent these finer-grained states of the inference program. 

\subsubsection{Detailed balance transitions with fixed state space}
\label{sec:application_db_fixed}

If we let $\mathcal{V}_t = \varnothing$ for $t=2,\ldots,T$, we recover a Markov chain
with fixed state space $\mathcal{V}_1 = \mathcal{U}_1 = \cdots = \mathcal{U}_{T-1} =
\mathcal{Z}$, and the inference program is the annealed importance sampling algorithm \cite{Neal2001}.
In this case, the estimated weight simplifies to
\begin{equation}
\hat{w}_y(z) = \frac{\tilde{p}_1(v_1)}{q(v_1;x^*)} \prod_{t=1}^{T-1} \frac{\tilde{p}_{t+1}(u_t)}{\tilde{p}_t(u_t)}
\end{equation}
Defining $u_0 := v_1$ and defining $p_0(u_0) := q(u_0;x^*)$, the estimated weight is:
\begin{equation}
\hat{w}_y(z) = \prod_{t=0}^{T-1} \frac{\tilde{p}_{t+1}(u_t)}{\tilde{p}_t(u_t)}
\end{equation}
Note that in this simplified setting, the approximating assumptions used to derive
the meta-inference distribution of Equation~\ref{eq:sai_metainference} are
$k_t(u';u) = p_t(u')$ for all $u, u'$ and $p_{t-1}(u) = p_t(u)$ for all $u$, for all $t=1,
\ldots, T$. The inference and meta-inference programs for this formulation are shown in
Algorithm~\ref{alg:db_fixed_inference} and
Algorithm~\ref{alg:db_fixed_metainference}.

\begin{algorithm}[H]
    \caption{Inference program for Section~\ref{sec:application_db_fixed}}
    \label{alg:db_fixed_inference}
    \begin{algorithmic}[1]
        \Require{Model program $p(z,x)$, dataset $x^*$, transition operators $k_1,\ldots,k_T$ satisfying
detailed balance with respect to $p_t$ where $p_T(z) = p(z|x^*)$, initializing distribution $p_0(z)$.}
        \Sample{$u_0$}{$p_0(z)$}
        \For{$t \gets 1 \textrm{ to } T-1$}
            \Sample{$u_t$}{$k_t(u;u_{t-1})$}
        \EndFor
        \Sample{$z$}{$k_T(z;u_{T-1})$}
        \State \Return{$(u_{0:T-1}, z)$}
    \end{algorithmic}
\end{algorithm}
\begin{algorithm}[H]
    \caption{Meta-inference program for Section~\ref{sec:application_db_fixed}}
    \label{alg:db_fixed_metainference}
    \begin{algorithmic}[1]
        \Require{Model program $p(z,x)$, dataset $x^*$, transition operators $k_1,\ldots,k_T$ satifying
detailed balance with respect to $p_t$ where $p_T(z) = p(z|x^*)$, initializing sample $z^*$.}
        \Sample{$u_{T-1}$}{$k_T(u;z^*)$}
        \For{$t \gets T-2 \textrm{ to } 0$}
            \Sample{$u_t$}{$k_{t+1}(u;u_{t+1})$}
        \EndFor
        \State \Return{$u_{0:T-1}$}
    \end{algorithmic}
\end{algorithm}

\subsubsection{Asymptotic gap between subjective divergence and symmetrized KL}
We now discuss how the quality of meta-inference is manifested in the subjective
divergence bounds for the sequential inference program defined in
Section~\ref{sec:application_db_fixed} and an oracle reference program. If we suppose that all transition
operators $k_t$ converge to their target distributions ($k_t(u';u) = p_t(u')
\forall u'$ for $t=1\ldots, T$), then the expected log estimated weight under the inference program
is:
\begin{align}
\E_{z\sim q(z;x^*)}\left[ \E_{y|z \sim q(y|z;x^*)} \left[ \log \hat{w}_y(z) \right] \right]
&=\E_{z \sim q(z;x^*)} \left[ \E_{y|z \sim q(y|z;x^*)} \left[ \sum_{t=0}^{T-1} \log \frac{\tilde{p}_{t+1}(u_t)}{\tilde{p}_t(u_t)} \right] \right]\\
&=\log p(x^*) + \sum_{t=0}^{T-1} \E_{u_t \sim p_t} \left[ \log \frac{p_{t+1}(u_t)}{p_t(u_t)}\right]\\
&=\log p(x^*) - \sum_{t=0}^{T-1} \KL(p_t(z)||p_{t+1}(z))
\end{align}
where we have used the fact that the normalizing constant of $\tilde{p}_T$ is
$p(x^*)$, that the normalizing constants of
$\tilde{p}_1,\ldots,\tilde{p}_{T-1}$ were arbitrary (and can be one), and that
$\tilde{p}_0$ is normalized. The expected log estimated weight under the reference program is:
\begin{align}
\E_{z\sim p(z|x^*)} \left[ \E_{y|z \sim m(y;z,x^*)} \left[ \log \hat{w}_y(z) \right] \right]
&=\E_{z \sim p(z|x^*)} \left[ \E_{y|z \sim m(y;z,x^*)} \left[ \sum_{t=0}^{T-1} \log \frac{\tilde{p}_{t+1}(u_t)}{\tilde{p}_t(u_t)} \right] \right]\\
&=\log p(x^*) + \sum_{t=0}^{T-1} \E_{u_t \sim p_{t+1}} \left[ \log \frac{p_{t+1}(u_t)}{p_t(u_t)}\right]\\
&=\log p(x^*) + \sum_{t=0}^{T-1} \KL(p_{t+1}(z)||p_t(z))
\end{align}
The subjective divergence with an oracle reference is the difference between these two expectations, which is the sum of 
symmetrized KL divergences between successive distributions in the sequence
$p_0(z), \ldots, p_T(z)$, where $p_T(z)$ is the posterior $p(z|x^*)$:
\begin{align} 
\SBJ(q(z;x^*)||p(z|x^*)) = \sum_{t=0}^{T-1} \KL(p_t(z)||p_{t+1}(z)) + \KL(p_{t+1}(z)||p_t(z)) \label{eq:sai_asymptotic_gap}
\end{align}
For inference programs for which the initialization distribution $p_0(u_0)$ is
the prior $p(z)$, this is the sum of symmetrized KL divergences between the
prior and the posterior of the inference problem. Note that in the limit of
convergence for each $k_t$ in the inference program, including $k_T$, the
approximating distribution equals the posterior ($q(z;x^*) = p(z|x^*)$) and the
true symmetrized KL divergence is zero. The gap between the asymptotic
subjective divergence of Equation~\ref{eq:sai_asymptotic_gap} and the
actual divergence of zero is a instance of the quantity defined in
Equation~\ref{eq:general_metainference_gap}, which quantifies the quality of
meta-inference. In this case, the asymptotic gap can be attributed to the
approximating assumption $p_{t-1}(z) = p_t(z)$ that was made when deriving the
meta-inference distribution.  

\subsubsection{Choice of target distribution sequence}

The asymptotic gap described in the previous section illustrates that the
subjective divergence profiles for this class of algorithms depends heavily on
the sequence of target distributions $p_t$. One generic sequence of target
distributions is the sequential observation sequence: $p_t(z) = p(z|y_{1:t})$.
The asymptotic subjective divergence bounds (Equation~\ref{eq:sai_asymptotic_gap}) for this sequence
depend on the data order.   

\subsubsection{Standard non-sequential MCMC}
We can represent the standard Markov chain Monte Carlo (MCMC) setting in which
a single target distribution $p(z|x^*)$ is targeted by a single kernel $k_1$
which satisfies detailed balance with respect to $p_1(z) = p(z|x^*)$ and is
composed of repeated application of primitive transition operators
which themselves satisfy detailed balance. In this case, the divergence bound
of Equation~\ref{eq:sai_asymptotic_gap} degenerates to the symmetrized KL divergence
between the initializing distribution $p_0(z)$ of the Markov chain and the
posterior, and no `credit' is given for running the transition operator. The
assumption $p_{t-1}(z) = p_t(z)$ used in deriving the meta-inference program
degenerates to $p_0(z) = p_1(z) = p(z|x^*)$, so the meta-inference program is of
low quality and the gap between the subjective divergence and the true symmetrized KL divergence
(given for the general case in Equation~\ref{eq:general_metainference_gap}) is large.

\subsubsection{Comparing convergence rates of transition operators}

Algorithm~\ref{alg:db_fixed_inference} and
Algorithm~\ref{alg:db_fixed_metainference}, combined with the subjective divergence estimation procedure of Algorithm~\ref{alg:general}, can be used as a test-bench for
subjectively comparing the convergence rates of transition operators.
Specifically, we instantiate sequential detailed balance inference programs
that utilize the same sequence of target distributions $p_t$, where we vary the
type of primitive transition operator used, and the number of consecutive
applications of the primitive transition operator within each of the $k_t$.
Note that the asymptotic subjective divergence
(Equation~\ref{eq:sai_asymptotic_gap}) is the same regardless of the type of
transition operators used within the $k_t$.

\subsection{Particle filtering}
Consider a state space model of the form
\begin{equation}
p(z,x) = p(z_{1:T}, x_{1:T}) = \left( p(z_1) \prod_{t=2}^T p(z_t | z_{t-1}) \right) \left( \prod_{t=1}^T p(x_t | z_t) \right)
\end{equation}
We apply the particle filter inference program as defined in \cite{holenstein2009particle},
Algorithm 2.3, with independent resampling, and derive
a meta-inference program that permits
Algorithm~\ref{alg:general} to be used to estimate subjective divergences of
this inference program with respect to the smoothing problem, with posterior $p(z_{1:T} | x_{1:T})$.

To simplify notation, we assume that a fixed number of particles $K$ is used at
each step of the particle filter. We denote the internal states of the particle
filter as $u^i_t$ for $i=1,\ldots,K$ and $t=1,\ldots,T$ and the internal
ancestor choices by $a^i_t$ for $i=1,\ldots,K$ and $t=1,\ldots,T-1$, where
$a^i_t$ is the index of the parent of state $u^i_{t+1}$, denoted
$u^{a^i_t}_{t}$. 
The full set of internal states is denoted $u^{1:K}_{1:T}$ and the full set of internal ancestor choices is denoted $a^{1:K}_{1:T-1}$.
The proposal densities are denoted $M_1(u_1)$ and
$M_t(u_t;u_{t-1})$ for $t=2\ldots,T$. An unnormalized weight is assigned to
each particle at each time step, for $i=1,\ldots,K$:
\begin{equation} \label{eq:pf_weight_1}
w^i_t := \frac{p(u^i_t|u^{a^i_{t-1}}_{t-1}) p(x_t|u^i_t)}{M_t(u^i_t;u^{a^i_{t-1}}_{t-1})}
\end{equation}
for $t=2,\ldots,T$, and
\begin{equation} \label{eq:pf_weight_2}
w^i_1 := \frac{p(u^i_1) p(x_1|u^i_1)}{M_1(u^i_1)}
\end{equation}
Note that these are not the same type of weight as the $w(z)$ used directly in the subjective divergence definition.
We assume that parent indices are
sampled independently from a categorical distribution given the normalized
weights. Conditioned on $u^{1:K}_{1:T}$ and $a^{1:K}_{1:T-1}$, a single final
particle index $k$ is sampled according to the normalized weights at the final
time step. A final hidden sequence $z_{1:T}$ is then generated
deterministically given $u_{1:T}^{1:K}$, $a^{1:K}_{1:T-1}$, and $k$ by selecting $z_t = u^{I_t}_t$ for $t=1,\ldots,T$
where $I_t$ is the ancestor index of state $u^k_T$ at time $t$,
defined recursively as $I_T := k$ and $I_t := a^{I_{t+1}}_t$ for $t=1,\ldots,T-1$.
We define the inference execution history
of the particle filter by $y = (u^{1:K}_{1:T}, a_{1:T-1}^{1:K},
k)$, and:
\begin{align}
q(y,z;x^*)
&= \left( \prod_{i=1}^K M_1(u_1^i) \right) \left( \prod_{t=2}^T \prod_{i=1}^K \frac{w^{a^i_{t-1}}_{t-1}}{\sum_{j=1}^K w^j_{t-1}} M_t(u_t^i; u_{t-1}^{a_{t-1}^i}) \right) \left( \frac{w_T^k}{\sum_{j=1}^K w_T^j} \right) \left( \prod_{t=1}^T \delta(u^{I_t}_t, z_t) \right)
\end{align}
For the meta-inference program $m(y;z,x^*)$, we use the conditional SMC (CSMC) update 
(\cite{holenstein2009particle}, Algorithm 3.3), which begins with a hidden state sequence
$z_{1:T}$ and its ancestry $I = (I_1, \ldots, I_T)$ and runs the particle
filter forward with this ancestry and particle states $u^{I_t}_t$ for $t=1,\ldots,T$ fixed.
Specifically, we first sample the ancestry $I$ uniformly at random:
($\mbox{Prob}(I) = \frac{1}{K^T}$), and then proceed with the CSMC update. The density of the meta-inference program is, assuming independent resampling in the particle filter:
\begin{align}
m(y;z,x^*)
&= \frac{1}{K^T} \left( \prod_{i \ne I_1} M_1(u_1^i) \right) \left( \prod_{t=2}^T \prod_{i \ne I_t} \frac{w^{a^i_{t-1}}_{t-1}}{\sum_{j=1}^K w^j_{t-1}} M_t(u_t^i; u_{t-1}^{a_{t-1}^i}) \right) \left( \prod_{t=1}^T \delta(u^{I_t}_t, z_t) \right)
\end{align}
The estimated weight then simplifies to:
\begin{align}
\hat{w}_y(z)
&= \frac{p(z,x^*) m(y;z,x^*)}{q(y,z;x^*)}\\
&= \frac{p(z,x^*) \frac{1}{K^T} \left( \prod_{i \ne I_1} M_1(u_1^i) \right) \left( \prod_{t=2}^T \prod_{i \ne I_t} \frac{w^{a^i_{t-1}}_{t-1}}{\sum_{j=1}^K w^j_{t-1}} M_t(u_t^i; u_{t-1}^{a_{t-1}^i}) \right) \left( \prod_{t=1}^T \delta(u^{I_t}_t, z_t) \right)}{\left( \prod_{i=1}^K M_1(u_1^i) \right) \left( \prod_{t=2}^T \prod_{i=1}^K \frac{w^{a^i_{t-1}}_{t-1}}{\sum_{j=1}^K w^j_{t-1}} M_t(u_t^i; u_{t-1}^{a_{t-1}^i}) \right) \left( \frac{w_T^k}{\sum_{j=1}^K w_T^j} \right) \left( \prod_{t=1}^T \delta(u^{I_t}_t, z_t) \right)} \\
\shortintertext{Ignoring $(y, z)$ for which $\prod_{t=1}^T \delta(u^{I_t}_t, z_t) = 0$ because these are not sampled under either $q(y,z;x^*)$ or $m(y;z,x^*)$:}
&= \frac{p(z,x^*) \frac{1}{K^T} \left( \prod_{i \ne I_1} M_1(u_1^i) \right) \left( \prod_{t=2}^T \prod_{i \ne I_t} \frac{w^{a^i_{t-1}}_{t-1}}{\sum_{j=1}^K w^j_{t-1}} M_t(u_t^i; u_{t-1}^{a_{t-1}^i}) \right)}{\left( \prod_{i=1}^K M_1(u_1^i) \right) \left( \prod_{t=2}^T \prod_{i=1}^K \frac{w^{a^i_{t-1}}_{t-1}}{\sum_{j=1}^K w^j_{t-1}} M_t(u_t^i; u_{t-1}^{a_{t-1}^i}) \right) \left( \frac{w_T^k}{\sum_{j=1}^K w_T^j} \right)} \\
\shortintertext{Canceling factors:}
&= \frac{1}{K^T} \frac{p(z,x^*)}{M_1(u_1^{I_1}) \left( \prod_{t=2}^T \frac{w^{a^{I_t}_{t-1}}_{t-1}}{\sum_{j=1}^K w^j_{t-1}} M_t(u_t^{I_t}; u_{t-1}^{a^{I_t}_{t-1}})\right) \left(\frac{w_T^k}{\sum_{j=1}^K w_T^j}\right)}\\
\shortintertext{Since $I_t := a^{I_{t+1}}_t$ for $t=1,\ldots,T-1$:}
&= \frac{1}{K^T} \frac{p(z,x^*)}{M_1(u_1^{I_1}) \left( \prod_{t=2}^T \frac{w^{I_{t-1}}_{t-1}}{\sum_{j=1}^K w^j_{t-1}} M_t(u_t^{I_t}; u_{t-1}^{I_{t-1}})\right) \left(\frac{w_T^k}{\sum_{j=1}^K w_T^j}\right)}\\
&= \left( \prod_{t=1}^T \frac{1}{K} \sum_{j=1}^K w^j_t \right) \left( \frac{p(z,x^*)}{M_1(u_1^{I_1}) \prod_{t=2}^T M_t(u_t^{I_t}; u_{t-1}^{I_{t-1}}) \prod_{t=1}^T w^{I_t}_t} \right)\\
\shortintertext{Using the definition of the particle filter's marginal likelihood estimate $\hat{Z} := \prod_{t=1}^T \frac{1}{K} \sum_{j=1}^K w^j_t$:}
&= \hat{Z} \left( \frac{p(z,x^*)}{M_1(u_1^{I_1}) \prod_{t=2}^T M_t(u_t^{I_t}; u_{t-1}^{I_{t-1}}) \prod_{t=1}^T w^{I_t}_t} \right)\\
\shortintertext{Expanding $p(z,x^*)$ and using the definitions of the weights of Equation~\ref{eq:pf_weight_1} and Equation~\ref{eq:pf_weight_2}:}
&= \hat{Z} \frac{p(z_1) \prod_{t=2}^T p(z_t | z_{t-1}) \prod_{t=1}^T p(x_t^* | z_t)}{p(u^{I_1}_1) \prod_{t=2}^T p(u^{I_t}_t|u^{I_{t-1}}_{t-1}) \prod_{t=1}^T p(x_t^* | u^{I_t}_t)}\\
\shortintertext{Using $z_t = u^{I_t}_t$ for $t=1,\ldots,T$:}
&= \hat{Z}
\end{align}

\subsubsection{Special case: sampling importance resampling (SIR)}
We can immediately apply the meta-inference program formulation for the particle
filter to non-state-space probabilistic models by considering the special case of
$T=1$. In this case, the weight estimate is
\begin{equation}
\hat{w}_y(z) = \frac{1}{K} \sum_{i=1}^K w^i_1 = \frac{1}{K} \sum_{i=1}^K \frac{p(u_1^i) p(x_1^* | u_1^i)}{M_1(u_1^i)}
\end{equation}
where $x_1^*$ contains all of the observations, and we recover sampling
importance resampling (SIR). The meta-inference program in this case places the
output $z$ into one of $K$ particles and samples the other $K-1$ particles from the
proposal distribution $M_1$.